\documentclass{article} 
\usepackage{iclr2027_conference,times}

\usepackage{amsmath,amsfonts,bm}

\def\eqref#1{equation~\ref{#1}}

\def\1{\bm{1}}

\DeclareMathAlphabet{\mathsfit}{\encodingdefault}{\sfdefault}{m}{sl}
\SetMathAlphabet{\mathsfit}{bold}{\encodingdefault}{\sfdefault}{bx}{n}

\usepackage{amsfonts}
\usepackage{amssymb}
\usepackage{array}
\usepackage{hyperref}
\usepackage{url}
\usepackage{booktabs}
\usepackage{multirow}
\usepackage{graphicx}
\usepackage{soul}
\usepackage{float}
\usepackage{colortbl}
\usepackage[ruled,vlined,linesnumbered]{algorithm2e}
\usepackage[T1]{fontenc}
\title{Just Initialize: A Training-Free Initialization Component for Large-Scale Routing Optimization}

\author{
\textbf{
Jiale Zhao\textsuperscript{1,*,$\ddagger$} \quad
Sirui Mao\textsuperscript{2,*} \quad
Zimu Chen\textsuperscript{1,*} \quad
Wentao Yang\textsuperscript{2,*}
}
\\[3pt]
\textbf{
Zihan Wang\textsuperscript{3} \quad
Xuefeng Huang\textsuperscript{4} \quad
Junji Cheng\textsuperscript{1} \quad
Liyuanjun Lai\textsuperscript{1,$\dagger$}
}
\\[2mm]
\textsuperscript{1}School of Automation Science and Electrical Engineering,
Beihang University, Beijing, China
\\
\textsuperscript{2}School of Computer Science and Engineering,
Beihang University, Beijing, China
\\
\textsuperscript{3}School of Cyber Science and Technology,
Beihang University, Beijing, China
\\
\textsuperscript{4}School of Mechanical Engineering and Automation,
Beihang University, Beijing, China
\\
\textbf{Contact:} \texttt{lailiyuanjun@buaa.edu.cn}
\\[1mm]
\textsuperscript{*}Equal contribution.
\quad
\textsuperscript{$\dagger$}Corresponding author.
\quad
\textsuperscript{$\ddagger$}Project lead.
}

\iclrfinalcopy 
\begin{document}

\maketitle

\begin{abstract}
Large-scale routing problems are difficult to solve efficiently as their search spaces grow rapidly with problem size. Existing approaches primarily improve the optimization procedure itself, often at increasing computational cost. We instead shift the focus to a useful initialization that can be refined into a high-quality solution with limited downstream refinement. We propose Just Initialize, a training-free and solver-agnostic initialization component for large-scale routing optimization. Just Initialize compresses a large routing instance into a compact surrogate space, optimizes its global routing structure, and recovers the resulting solution as an optimization-friendly starting point in the original space. Extensive experiments on Traveling Salesman Problems (TSPs), Capacitated Vehicle Routing Problems (CVRPs), Vehicle Routing Problems with Time Windows (VRPTWs), and Prize-Collecting Traveling Salesman Problems (PCTSPs) demonstrate that Just Initialize achieves high-quality solutions comparable to or better than state-of-the-art methods while substantially reducing computational cost across instances ranging from 1K to 100K nodes, including an average speedup of approximately 70$\times$, sub-second runtimes on 10K-node instances, and runtimes within tens of seconds on 100K-node instances. The source code is available at https://github.com/BeihangAILab/Just-Initialize

\end{abstract}

\section{Introduction}
Routing problems are widely encountered in real-world systems such as transportation, autonomous delivery, and industrial workshops~\citep{golden2023evolution,elshaer2020taxonomic}. 
However, their NP-hard nature makes solving large-scale instances computationally expensive and time-consuming. 
Traditional optimization methods have achieved remarkable success in obtaining optimal or near-optimal solutions. 
Deterministic algorithms, such as mixed integer programming algorithms, provide strong optimality guarantees but suffer from prohibitive computational costs as problem scales increase. 
Heuristics and metaheuristics improve search scalability but often rely on iterative evolution and evaluation~\citep{yang2021branch,wang2021deep}. 
These limitations significantly hinder their applicability to large-scale real-world scenarios, where efficient decision making is increasingly required.

Recently, learning-based methods have emerged as promising alternatives to handcrafted routing solvers. Neural combinatorial optimization (NCO) uses neural networks to learn solution construction or improvement strategies through various training paradigms. Among these, reinforcement learning (RL) trains routing policies through reward-driven interactions with problem environments~\citep{kwon2020pomo,kim2021learning}. Although NCO methods enable efficient inference on unseen instances~\citep{bi2022learning,luo2023neural}, scaling them to large routing problems remains challenging. Recent studies address this challenge through enhanced neural solvers, hierarchical decomposition, search space reduction, and neural-guided optimization~\citep{ye2024glop,li2021learning,qiu2022dimes}. Despite substantial progress, these approaches primarily improve the solving process itself, often introducing increasingly sophisticated architectures, training procedures, or search mechanisms. This motivates a complementary perspective: improving where optimization begins.

Rather than designing more sophisticated optimization procedures, we ask whether the computational burden can be reduced by improving where optimization starts. A good initialization should therefore be judged not only by its immediate solution quality, but also by how efficiently it can be refined into a high-quality solution. A solution with a better initial objective value may require substantial further search, while one with a worse initial value may reach a better result through limited refinement. This perspective shifts the goal of initialization toward identifying starting points with strong refinement potential. Since such starting points are primarily characterized by their global structure rather than fully resolved local decisions, compression provides a natural way to reduce the search space while retaining the information most relevant to subsequent refinement.

Previous studies have explored compression and coarsening to reduce problem size before recovering solutions in the original space~\citep{orloff1976reduction,nalecz2025graph}. When recovery is expected to produce a high-quality solution directly, the compressed representation must preserve both global structure and local details, limiting how aggressively the problem can be compressed. 

Therefore, we propose Just Initialize, a training-free and solver-agnostic initialization component for large-scale routing optimization. Rather than directly searching for a final solution in the original space, Just Initialize compresses a large routing instance into a compact surrogate space, where promising global structures can be identified efficiently. The resulting compact solution is then recovered as an optimization-friendly starting point in the original space, leaving fine-grained decisions to downstream refinement. In this way, Just Initialize focuses on discovering where optimization should begin rather than resolving the entire solution from scratch.
\begin{figure}[t]
\centering
\begin{minipage}[b]{0.62\textwidth}
    \centering
    \includegraphics[width=\textwidth]{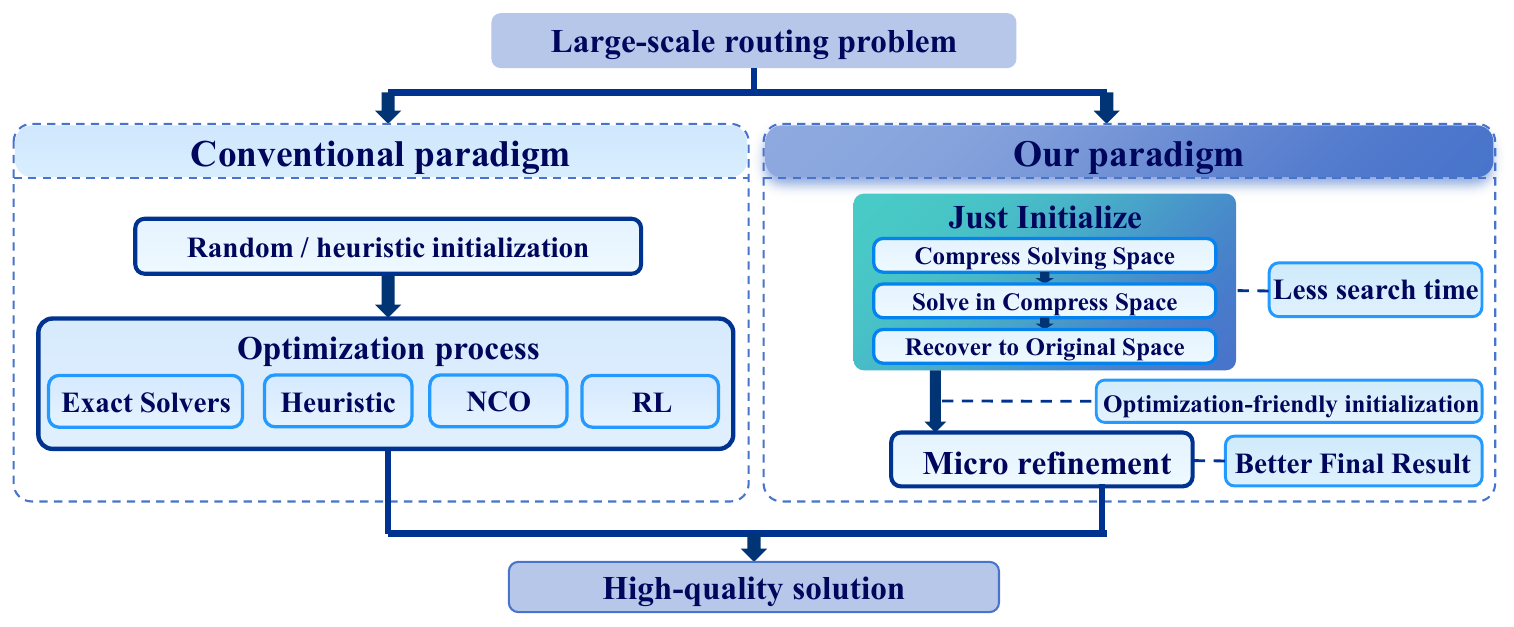}
    \par\vspace{2pt}\textbf{(a)}
\end{minipage}
\hfill
\begin{minipage}[b]{0.36\textwidth}
    \centering
    \includegraphics[width=\textwidth]{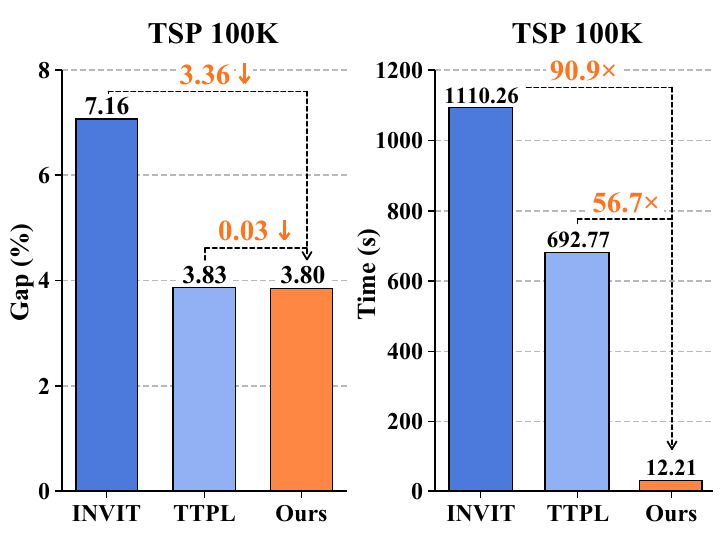}
    \par\vspace{2pt}\textbf{(b)}
\end{minipage}

\caption{Rethinking large-scale routing optimization.
(a) Conventional paradigm and our paradigm.
(b) Performance comparison with SOTA methods on 100K-node TSP instances.}
\label{fig:overview}
\end{figure}

Our contributions are summarized as follows:

\begin{itemize}
   \item We introduce \textbf{Just Initialize}, a training-free and solver-independent initialization component that discovers optimization-friendly starting points for large-scale routing problems by separating solution discovery from solution optimization.
   
   \item We develop a geometry-guided instance compression procedure to construct compact routing representations, and propose an initialization-oriented objective to evaluate compact solutions according to their downstream optimization potential.
   
   \item Extensive experiments on TSPs, CVRPs, VRPTWs, and PCTSPs demonstrate that Just Initialize achieves competitive or superior solution quality compared with state-of-the-art methods, while substantially reducing computational cost and scaling to instances from 1K to 100K nodes.

\end{itemize}

\section{Related Works}
\subsection{Traditional Optimization Methods}

Traditional routing problems are commonly addressed by exact algorithms, heuristics, and metaheuristics. Exact methods provide strong optimality guarantees but become increasingly expensive as problem size grows~\citep{gamst2024exact,aertsveenstra2024unified}. For large-scale instances, heuristic and metaheuristic solvers instead rely on carefully designed neighborhoods and iterative search, including Lin--Kernighan--Helsgaun~\citep{helsgaun2000effective}, adaptive large neighborhood search~\citep{ropke2006adaptive}, hybrid genetic search~\citep{vidal2022hgs,simensen2022combining}, and specialized local-search procedures~\citep{cook2024constrained}. Recent methods further demonstrate that carefully engineered search can scale to extremely large routing instances~\citep{accorsi2024routing}. Despite their strong performance, obtaining high-quality solutions typically still requires substantial search as problem size increases.

\subsection{Neural Combinatorial Optimization}

Neural combinatorial optimization (NCO) learns solution construction or improvement strategies directly from problem instances~\citep{bengio2021machine}. Reinforcement learning is widely used for training such policies, with representative methods including Attention Model~\citep{kool2019attention}, POMO~\citep{kwon2020pomo}, collaborative policies~\citep{kim2021learning}, knowledge distillation~\citep{bi2022learning}, and symmetry-aware learning~\citep{kim2022symnco}. These methods reduce dependence on manually designed search rules and enable efficient inference after training.

Recent work has increasingly focused on scaling NCO to large routing instances. Existing approaches improve model architectures and training strategies~\citep{luo2023neural,fang2024invit,luo2025boosting,chen2025ttpl,luo2025insert}, decompose large problems into smaller subproblems~\citep{li2021learning,pan2023htsp,ye2024glop,zheng2024udc}, learn alternative solution representations~\citep{qiu2022dimes,sun2023difusco}, or combine neural predictions with classical optimization procedures~\citep{xin2021neurolkh,kool2022dpdp,ye2023deepaco}. These methods improve scalability from different perspectives, but primarily focus on how solutions are constructed, represented, or optimized during the search process.

\subsection{Problem Reduction and Compression}

Problem reduction provides another way to improve routing scalability by reducing the number of decisions exposed to the solver. Classical approaches include cluster-first and route-first strategies~\citep{gillett1974heuristic,fisher1981generalized,beasley1983route,prins2014orderfirst}, explicit problem reduction~\citep{orloff1976reduction}, restricted neighborhood search~\citep{toth2003granular}, and subproblem-based optimization~\citep{queiroga2021popmusic}. More recent studies extend this idea through large-scale decomposition~\citep{kerscher2025decompose} and graph coarsening, where a smaller routing representation is constructed and subsequently expanded to the original problem~\citep{nalecz2025graph}.

Different from existing reduction methods that aim to preserve sufficient information for solution recovery, we use compression to identify optimization-friendly starting points, while leaving detailed routing decisions to subsequent optimization.

\section{Preliminaries}
\label{sec:preliminaries}

\subsection{Routing Problems and Solution Spaces}
\label{sec:routing_spaces}

Let $\mathcal{I}$ be the space of routing instances. Each instance 
$I=(G,X,K)$ defines a feasible solution space $\Omega(I)$ 
and an objective function $C_I$. To provide a unified representation 
across different routing variants, we represent a multi-route solution 
as a single visiting sequence with route boundaries. Specifically, a 
solution with routes $(r_1,\ldots,r_m)$ is written as

\begin{equation}
\pi=(\sigma,\mathbf{b}),
\qquad
\sigma=r_1\oplus\cdots\oplus r_m,
\qquad
b_j=\sum_{\ell=1}^{j}|r_\ell|,
\quad j=1,\ldots,m ,
\end{equation}

where $\sigma$ denotes the visiting sequence and $\mathbf{b}$ specifies 
the breakpoints that partition the sequence into individual routes. 
Given $\mathbf{b}$, each route can be recovered from the corresponding 
segment of $\sigma$. This representation is equivalent to the original 
multi-route formulation, while allowing different routing problems to 
share a common solution representation. For example, a solution to either TSP or PCTSP contains a single segment, whereas CVRP and VRPTW impose additional 
constraints on route segments and visited nodes.

The optimization objective is defined as

\begin{equation}
\pi^{\mathrm{*}}
\in
\arg\min_{\pi\in\Omega(I)} C_I(\pi).
\end{equation}

For later refinement, we define a local neighborhood over the original 
solution space. A neighboring solution is generated by modifying either 
the visiting sequence or the route structure while preserving feasibility:

\begin{equation}
N_I(\pi)
=
\left\{
(T_{i,j}(\sigma),\mathbf{b})
:
T_{i,j}\in
\{\mathrm{swap}_{i,j},\mathrm{rev}_{i:j}\},
i<j
\right\}
\cap \Omega(I),
\end{equation}

where $\mathrm{swap}_{i,j}$ exchanges two positions in the visiting 
sequence and $\mathrm{rev}_{i:j}$ reverses the segment between positions 
$i$ and $j$. This neighborhood defines the local modifications used by 
downstream refinement procedures.

\subsection{Compact Representations and Solution Space Mapping}
\label{sec:Compact_Representations}
Given an original routing instance $I$, we first construct a compact instance by grouping original nodes into a small number of compact units.
\begin{equation}
\tilde{I}=\Phi_c(I).
\end{equation}
The compact instance $\tilde I$ induces a lower-dimensional solution
space $\Omega(\tilde I)$, where each compact unit represents a group
of original nodes. This correspondence defines a projection from
the original solution space to the compact solution space:
\begin{equation}
\tilde{\pi}=\Phi_s(\pi;M),
\qquad
\pi\in\Omega(I),\quad
\tilde{\pi}\in\Omega(\tilde I).
\end{equation}

Here, $M$ records the correspondence between original nodes and compact units established during instance compression. The projection $\Phi_s(\cdot)$ does not directly compress routing decisions. Instead, it preserves the coarse structural information induced by instance compression while omitting fine-grained decisions. Therefore, the projection from the original solution space to the compact solution space is generally
many-to-one, where multiple original solutions may correspond to the same compact solution:
\begin{equation}
\Phi_s^{-1}(\tilde{\pi};M)
=
\{\pi\in\Omega(I)|\Phi_s(\pi;M)=\tilde{\pi}\}.
\end{equation}

\subsection{Initialization Quality Measurement }
\label{sec:refinement_initialization}

The quality of an initialization is determined by how effectively it enables subsequent optimization, rather than solely by its immediate objective value. A feasible solution $\pi_0\in\Omega(I)$ is regarded as an initialization
when it is provided as the starting point for downstream optimization. Let $\mathcal{R}_I(\pi_0;B)$ denote the solution obtained by applying a downstream solver initialized at $\pi_0$ with computational budget $B$. The budget can be measured in runtime, search iterations, or objective evaluations. We first evaluate an initialization by the expected objective after optimization:
\begin{equation}
J_B(\pi_0;I,\mathcal{R})
=
\mathbb{E}
\left[
C_I(\mathcal{R}_I(\pi_0;B))
\right],
\end{equation}
where the expectation accounts for randomness in stochastic solvers and can be omitted for deterministic procedures. Because $C_I$ is a cost to be minimized, a lower $J_B$ indicates a more
effective initialization, and initializations are compared by minimizing $J_B$
under a fixed solver and budget.

\section{Methodology}
\label{sec:methodology}

\subsection{Overview of Just Initialize}
\label{sec:overview}
Building on the formulation in Section~\ref{sec:preliminaries}, Just Initialize constructs an optimization-friendly initialization through three stages: instance
compression, compact-space optimization, and solution recovery.

The central idea is to separate solution discovery from solution optimization. Instead of directly searching in the original high-dimensional solution space, Just Initialize first compresses the problem into a compact space, where coarse routing structures can be efficiently explored. The obtained compact solution is then recovered into the original space as an optimization-friendly initialization and further improved through downstream refinement. The following sections describe each component in detail.

\begin{figure}[t]
    \centering
    \includegraphics[width=\linewidth]{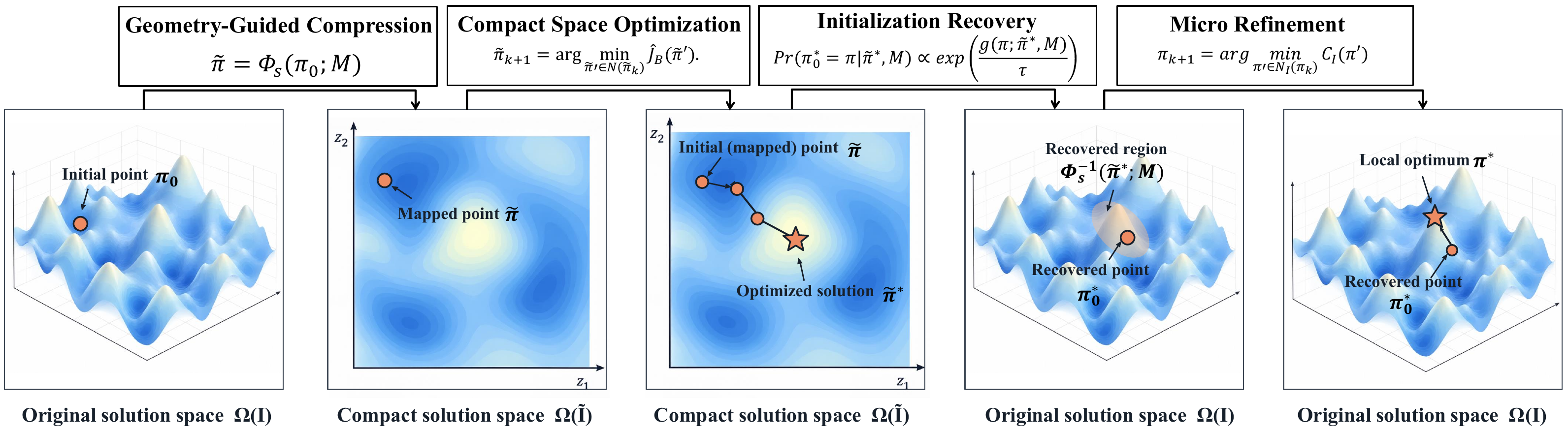}
    \caption{
    Overview of Just Initialize. The framework compresses the original solution space, optimizes in a compact space, recovers an optimization-friendly initialization, and performs lightweight refinement in the original space.
    }
    \label{fig:solution_space_mapping}
\end{figure}

\subsection{Geometry-Guided Instance Compression for Solution Space Reduction}
The first stage constructs a compact representation of the original routing
instance. Different from conventional compression methods that aim to recover
a high-quality solution directly, our goal is to identify optimization-friendly
regions while leaving fine-grained decisions to subsequent refinement.

Given an original instance $I$, we construct a compact instance by grouping
nodes into compact units:
$\tilde I=\Phi_c(I).$
The compact instance induces a reduced solution space $\Omega(\tilde I)$,
where solutions are defined over compact units instead of individual nodes.
Accordingly, the original solution space is projected through
\begin{equation}
\tilde{\pi}=\Phi_s(\pi;M),
\quad
\pi\in\Omega(I),\tilde{\pi}\in\Omega(\tilde I).
\end{equation}
Since multiple original solutions may share the same compact structure,
$\Phi_s$ is generally many-to-one. The compact representation preserves coarse geometric structures and boundary
information required for recovery, while omitting unresolved local decisions. The detailed
construction procedure is summarized in Algorithm~\ref{alg:compression}.

\begin{algorithm}
\caption{Geometry-Guided Instance Compression for Solution Space Reduction}
\label{alg:compression}
\KwIn{Routing instance $I=(G,X,K)$; target unit size $b$}
\KwOut{Compact instance $\tilde I$, correspondence information $M$, and solution projection $\Phi_s$.}

$F \leftarrow \textsc{ExtractGeometry}(G,X)$

$N \leftarrow \textsc{BuildNeighborhood}(V,F)$

$C \leftarrow \textsc{ConstructCompactUnits}(V,N,F,b,K)$

\ForEach{$C_i\in C$}{
    $z_i \leftarrow \textsc{SummarizeGeometry}(C_i,F)$\;
    
    $\mathcal T_i \leftarrow \textsc{ExtractBoundaryStates}(C_i,F,K)$\;
    
    $M_i\leftarrow(C_i,z_i,\mathcal T_i,F|_{C_i})$\;
}

$\tilde C\leftarrow\textsc{EstimateConnections}(C,\{z_i,\mathcal T_i\},I)$

$\tilde K\leftarrow\textsc{AggregateConstraints}(C,K)$

$\tilde I\leftarrow(C,\{z_i,\mathcal T_i\},\tilde C,\tilde K)$

$\Phi_s \leftarrow \textsc{ConstructSolutionProjection}(C,M)$

\Return $(\tilde I,M,\Phi_s)$
\end{algorithm}

\subsection{Solving in the Compact Space}
After compression, the routing problem is optimized in the compact solution space $\Omega(\tilde I)$, where each solution represents the ordering of compact units and their corresponding local states. Following the initialization quality
criterion defined in Section~\ref{sec:refinement_initialization}, we evaluate a compact solution according to the downstream performance of its recovered initialization.

Since $J_B$ is defined as an expectation over downstream optimization processes, its exact value is generally unavailable. We therefore estimate it using Monte Carlo sampling with a small number of refinement trials:
\begin{equation}
\hat{J}_B(\tilde{\pi})
=
\frac{1}{S}
\sum_{s=1}^{S}
C_I(R_I^{(s)}(\pi_0;B)),
\end{equation}
where $S$ denotes the number of sampled refinement trajectories and $R_I^{(s)}$ represents the $s$-th refinement process.

The compact-space search then aims to identify solutions with better initialization potential. Starting from an initial compact solution, we iteratively update the solution
by minimizing $\hat J_B$, the estimated initialization quality. Similar to the neighborhood operators defined in the
original solution space, the compact-space neighborhood $N(\tilde{\pi})$ is constructed by applying local modifications to compact-unit ordering while
preserving compact feasibility:
\begin{equation}
\tilde{\pi}_{k+1}
=
\arg\min_{\tilde{\pi}'\in N(\tilde{\pi}_k)}
\hat{J}_B(\tilde{\pi}').
\end{equation}
The final compact solution $\tilde{\pi}^{*}$ captures a global routing structure with strong refinement potential and is subsequently recovered to generate an optimization-friendly initialization in the original solution space.

\subsection{Recovering Optimization-Friendly Initializations}

Given the final compact solution $\tilde{\pi}^{*}$, recovery selects an
original-space initialization from the feasible region
$\Phi_s^{-1}(\tilde{\pi}^{*};M)$ induced by the solution-space mapping defined
in Section~\ref{sec:Compact_Representations}. The structural information $M$ extracted during compression provides guidance for evaluating the compatibility between candidate solutions
and the compact representation. Specifically, we define a structural consistency score
$g(\pi;\tilde{\pi}^{*},M)$ to measure how well an original-space solution preserves the compact-unit ordering, geometric structure, and boundary configurations specified by $\tilde{\pi}^{*}$ and $M$. The detailed formulation of $g(\pi;\tilde{\pi}^{*},M)$ is provided in Appendix~\ref{app:implementation}. The initialization is selected according to:

\begin{equation}
\Pr(\pi_0^{*}=\pi \mid \tilde{\pi}^{*},M)
=
\frac{
\exp(g(\pi;\tilde{\pi}^{*},M)/\tau)
}{
\sum_{\pi'\in\Phi_s^{-1}(\tilde{\pi}^{*};M)}
\exp(g(\pi';\tilde{\pi}^{*},M)/\tau)
},
\quad
\pi\in\Phi_s^{-1}(\tilde{\pi}^{*};M).
\end{equation}

Here, $\tau$ controls the trade-off between selecting highly compatible candidates and maintaining diversity. The resulting optimization friendly initialization $\pi_0^{*}$ is then passed to downstream refinement in the original solution space.

\subsection{Micro refinement}
\label{sec:downstream_refinement}

The $\pi_0^{*}$ serves as an optimization-friendly initialization rather than a final solution. We therefore perform a lightweight local refinement in the original solution space using the neighborhood operators defined in Section~\ref{sec:routing_spaces}.

At each iteration, we select the best improving neighbor:
\[
\pi_{k+1}
=
\arg\min_{\pi'\in N_I(\pi_k)}
C_I(\pi'),
\quad
C_I(\pi_{k+1})<C_I(\pi_k).
\]
If no improving neighbor exists, the procedure terminates. Since the initialization already captures the global
routing structure, this lightweight refinement only needs to correct remaining local decisions without rebuilding the solution.

\section{Experiment}
\begin{table*}[t]
\centering
\caption{
Comparison results on large-scale TSP and CVRP instances.
Obj. and Gap denote the solution objective and relative gap, respectively.
Bold marks the lowest Gap and Time at each scale before rounding; gray highlights our method. OOM: results that exceed GPU memory limits. OOT: the method fails to finish within the predefined time limit.
}
\label{tab:tsp_cvrp_main}

\resizebox{\textwidth}{!}{
\begin{tabular}{l|ccc|ccc|ccc|ccc|ccc}
\toprule
\multirow{2}{*}{\textbf{Method}}
& \multicolumn{3}{c|}{\textbf{TSP5K}}
& \multicolumn{3}{c|}{\textbf{TSP10K}}
& \multicolumn{3}{c|}{\textbf{TSP20K}}
& \multicolumn{3}{c|}{\textbf{TSP50K}}
& \multicolumn{3}{c}{\textbf{TSP100K}} \\
& Obj.$\downarrow$ & Gap & Time
& Obj.$\downarrow$ & Gap & Time
& Obj.$\downarrow$ & Gap & Time
& Obj.$\downarrow$ & Gap & Time
& Obj.$\downarrow$ & Gap & Time \\
\midrule

LKH-3 (Reference)
& 50.88 & -- & --
& 71.76 & -- & --
& 101.39 & -- & --
& 159.98 & -- & --
& 225.95 & -- & -- \\

\midrule

POMO (no aug.)
& 88.33 & 73.62 & 50.6\,s
& 133.64 & 86.23 & 6.5\,min
& \multicolumn{3}{c|}{OOM}
& \multicolumn{3}{c|}{OOM}
& \multicolumn{3}{c}{OOM} \\

POMO ($\times 8$ aug.)
& 87.62 & 72.23 & 7.7\,min
& \multicolumn{3}{c|}{OOM}
& \multicolumn{3}{c|}{OOM}
& \multicolumn{3}{c|}{OOM}
& \multicolumn{3}{c}{OOM} \\

LEHD (greedy)
& 59.50 & 16.94 & 1.2\,min
& 90.26 & 25.77 & 7.9\,min
& \multicolumn{3}{c|}{OOM}
& \multicolumn{3}{c|}{OOM}
& \multicolumn{3}{c}{OOM} \\

INViT
& 54.22 & 6.57 & 46.1\,s
& 76.76 & 6.97 & 1.5\,min
& 108.57 & 7.08 & 3.1\,min
& 171.42 & 7.15 & 8.3\,min
& 242.13 & 7.16 & 18.5\,min \\

SIL
& 57.23 & 12.49 & 1.4\,min
& 93.50 & 30.29 & 10.5\,min
& 179.04 & 76.59 & 1.6\,h
& \multicolumn{3}{c|}{OOM}
& \multicolumn{3}{c}{OOM} \\

L2C-Insert ($I=1000$)
& 51.73 & \textbf{1.68} & 57.1\,s
& 74.30 & 3.54 & 58.4\,s
& 105.15 & 3.71 & 1.0\,min
& 166.13 & 3.84 & 1.2\,min
& 237.29 & 5.02 & 1.6\,min \\

\midrule

DIFUSCO
& 52.48 & 3.15 & 26.0\,s
& 73.97 & \textbf{3.07} & 1.3\,min
& 105.35 & 3.91 & 4.5\,min
& 166.03 & 3.78 & 29.7\,min
& \multicolumn{3}{c}{OOM} \\

\midrule

GLOP
& 54.12 & 6.38 & 1.5\,s
& 76.59 & 6.73 & 1.8\,s
& 108.25 & 6.76 & 3.3\,s
& 170.91 & 6.83 & 17.0\,s
& 241.53 & 6.90 & 1.2\,min \\

UDC
& 52.82 & 3.83 & 3.5\,s
& 74.69 & 4.08 & 3.8\,s
& 105.66 & 4.21 & 5.2\,s
& 166.87 & 4.30 & 18.7\,s
& 235.84 & 4.38 & 1.2\,min \\

H-TSP
& 55.07 & 8.24 & 5.1\,s
& 77.85 & 8.48 & 9.3\,s
& 110.18 & 8.67 & 18.8\,s
& 173.89 & 8.69 & 47.3\,s
& 245.50 & 8.65 & 1.7\,min \\

\midrule

TTPL
& 52.81 & 3.79 & 33.6\,s
& 74.75 & 4.17 & 1.1\,min
& 105.51 & 4.06 & 2.3\,min
& 166.31 & 3.95 & 5.8\,min
& 234.61 & 3.83 & 11.5\,min \\

\midrule
JI+VND
& 53.35 & 4.87 & 15.3\,s
& 75.24 & 4.85 & 28.9\,s
& 106.63 & 5.16 & 31.5\,s
& 168.14 & 5.10 & 1.7\,min
& 238.92 & 5.74 & 2.7\,min \\
\rowcolor{gray!35}
JI+Refine (Ours)
& 52.56 & 3.30 & \textbf{0.24\,s}
& 74.20 & 3.40 & \textbf{0.53\,s}
& 105.00 & \textbf{3.56} & \textbf{1.47\,s}
& 165.77 & \textbf{3.62} & \textbf{4.83\,s}
& 234.20 & \textbf{3.65} & \textbf{12.21\,s} \\

\midrule
\midrule


\multirow{2}{*}{\textbf{Method}}
& \multicolumn{3}{c|}{\textbf{CVRP5K}}
& \multicolumn{3}{c|}{\textbf{CVRP10K}}
& \multicolumn{3}{c|}{\textbf{CVRP20K}}
& \multicolumn{3}{c|}{\textbf{CVRP50K}}
& \multicolumn{3}{c}{\textbf{CVRP100K}} \\
& Obj.$\downarrow$ & Gap & Time
& Obj.$\downarrow$ & Gap & Time
& Obj.$\downarrow$ & Gap & Time
& Obj.$\downarrow$ & Gap & Time
& Obj.$\downarrow$ & Gap & Time \\
\midrule

LKH-3 (Reference)
& 86.00 & -- & --
& 97.97 & -- & --
& 125.10 & -- & --
& 236.46 & -- & --
& 409.25 & -- & -- \\

OR-Tools
& 108.10 & 25.69 & 2.0\,min
& 131.90 & 34.64 & 2.0\,min
& \multicolumn{3}{c|}{OOM}
& \multicolumn{3}{c|}{OOM}
& \multicolumn{3}{c}{OOM} \\

\midrule

LEHD
& 104.51 & 21.52 & 1.4\,min
& \multicolumn{3}{c|}{OOM}
& \multicolumn{3}{c|}{OOM}
& \multicolumn{3}{c|}{OOM}
& \multicolumn{3}{c}{OOM} \\

INViT
& 104.53 & 21.55 & 1.8\,min
& 130.29 & 32.99 & 3.8\,min
& 183.17 & 46.42 & 11.4\,min
& 355.78 & 50.46 & 57.4\,min
& 604.34 & 47.67 & 3.6\,h \\

SIL
& 103.24 & 20.05 & 2.6\,min
& 127.60 & 30.24 & 5.4\,min
& 191.14 & 52.79 & 11.4\,min
& \multicolumn{3}{c|}{OOM}
& \multicolumn{3}{c}{OOM} \\

L2C-Insert
& 114.46 & 33.08 & 59.7\,s
& 148.95 & 52.04 & 2.0\,min
& 216.71 & 73.24 & 4.0\,min
& 409.53 & 73.19 & 10.0\,min
& 680.13 & 66.19 & 20.1\,min \\

\midrule

GLOP
& 122.28 & 42.19 & 3.1\,s
& 129.20 & 31.88 & 10.2\,s
& 174.14 & 39.20 & 27.3\,s
& \multicolumn{3}{c|}{OOM}
& \multicolumn{3}{c}{OOM} \\

UDC
& 155.31 & 80.59 & 14.9\,s
& 361.79 & 269.29 & 17.2\,s
& 534.78 & 327.48 & 27.5\,s
& \multicolumn{3}{c}{OOM}
& \multicolumn{3}{c}{OOM} \\

\midrule

TTPL
& 101.81 & 18.38 & 44.6\,s
& 127.61 & 30.25 & 1.5\,min
& 179.09 & 43.16 & 3.0\,min
& 345.87 & 46.27 & 7.5\,min
& 606.71 & 48.25 & 15.1\,min \\

\midrule

\rowcolor{gray!35}
JI+Refine (Ours)
& 99.32 & \textbf{15.56} & \textbf{0.43\,s}
& 117.31 & \textbf{19.80} & \textbf{0.9\,s}
& 153.62 & \textbf{22.77} & \textbf{2.2\,s}
& 284.74 & \textbf{20.51} & \textbf{7.9\,s}
& 474.09 & \textbf{16.05} & \textbf{18.9\,s} \\

\bottomrule
\end{tabular}
}
\end{table*}
\subsection{Experiment Settings}
\label{sec:experiment_settings}

\textbf{Datasets.}
We evaluate Just Initialize on four representative routing problems:
TSP, CVRP, VRPTW, and PCTSP, covering large-scale instances from 1K to 100K nodes.
For TSP and CVRP, we follow previous large-scale synthetic benchmarks and use 80 instances per problem, including 16 instances at each scale of 5K, 10K, 20K, 50K, and 100K nodes.
For VRPTW and PCTSP, we use 56 instances for each problem following established generation protocols.

\textbf{Baselines.}
We compare against representative methods from classical optimization,
neural constructive optimization, heatmap-based approaches,
decomposition methods, and neural-enhanced metaheuristics.
Empirical reference solutions are generated independently using LKH-3 for TSPs and CVRPs, HGS-PyVRP for VRPTWs, and LKH-3-PCTSP for PCTSPs.
These references are used only for computing relative gaps.

\textbf{Metrics.}
We report objective value, relative gap, and time cost.
For Just Initialize, runtime includes compression, compact-space optimization,
solution recovery, and downstream refinement.
We additionally evaluate initialization effectiveness by comparing downstream optimization from Just Initialize with alternative starting points.

\paragraph{Implementation Details.}
 Experiments are conducted on an Intel Xeon Platinum 8358P 64-core CPU, eight
NVIDIA GeForce RTX 4090 GPUs, and 256 GB RAM. Each configuration is run three times, and the results are averaged; objective values and gaps are reported only for feasible solutions.
\subsection{Comparative Study}
The results are summarized in Tables~\ref{tab:tsp_cvrp_main} and~\ref{tab:large_scale_variants}.
Across four routing benchmarks, \textbf{JI+Refine} consistently achieves competitive or superior solution quality while substantially reducing computational cost. On large-scale TSP and CVRP instances, JI+Refine scales to 100K nodes and maintains stable solution gaps, while many existing approaches suffer from rapidly increasing runtime or fail to complete due to computational limitations. Notably, JI+VND also achieves competitive performance compared with many NCO-based methods, demonstrating that the effectiveness of JI mainly comes from providing better optimization-friendly initializations rather than relying on a specific refinement strategy. For VRPTW and PCTSP, JI+Refine achieves competitive solution quality across different scales while reducing the solving time from minutes or hours to seconds.

Overall, JI+Refine achieves an average speedup of approximately 70$\times$ compared with existing approaches while preserving comparable solution quality. On 10K-node instances, JI+Refine completes optimization within seconds, and even 100K-node instances can be solved within tens of seconds. These results demonstrate that optimization-friendly initialization, combined with effective refinement, provides an efficient way to reduce the computational burden of large-scale routing optimization.

Beyond the large-scale synthetic benchmarks, we further evaluate JI+Refine under diverse instance distributions, including non-uniform TSP settings with clustered, explosion, and implosion patterns. We also test JI+Refine on widely used TSPLIB and CVRPLIB benchmarks to examine its generalization ability beyond synthetic settings. Across these additional benchmarks, JI+Refine maintains competitive solution quality with consistent computational efficiency, demonstrating robustness across different spatial distributions and benchmark sources.
Detailed results are provided in Appendix~\ref{app:Extended-Experiments}.

\begin{table}[H]
\centering
\caption{Comparison results on large-scale VRPTW and PCTSP instances.}
\label{tab:large_scale_variants}

\resizebox{\textwidth}{!}{
\begin{tabular}{l|ccc|ccc|ccc|ccc}
\toprule
Method
& \multicolumn{3}{c|}{VRPTW2K}
& \multicolumn{3}{c|}{VRPTW5K}
& \multicolumn{3}{c|}{VRPTW10K}
& \multicolumn{3}{c}{VRPTW20K}
\\

& Obj. $\downarrow$ & Gap & Time
& Obj. $\downarrow$ & Gap & Time
& Obj. $\downarrow$ & Gap & Time
& Obj. $\downarrow$ & Gap & Time
\\
\midrule

HGS-PyVRP (Reference)
& 335.93 & - & -
& 823.85 & - & -
& 1419.13 & - & -
& 2479.68 & - & - \\

OR-Tools
& 352.83 & 5.03 & 30.0\,min
& 891.49 & 8.21 & 2.0\,min
& 1498.60 & 5.60 & 10.0\,min
& 2584.07 & 4.21 & 30.0\,min \\

VROOM
& 342.08 & \textbf{1.83} & 6.0\,min
& 831.43 & \textbf{0.92} & 14.7\,min
& - & OOM & -
& - & OOM & - \\

ALNS
& 381.58 & 13.59 & 3.8\,h
& 901.13 & 9.38 & 3.6\,h
& 1520.17 & 7.12 & 3.1\,h
& 2615.07 & 5.46 & 3.3\,h \\

\rowcolor{gray!35}
JI+Refine (Ours)
& 362.91 & 8.25 & \textbf{2.9\,s}
& 863.80 & 5.16 & \textbf{9.6\,s}
& 1478.16 & \textbf{4.43} & \textbf{9.3\,s}
& 2574.90 & \textbf{3.96} & \textbf{15.8\,s} \\

\midrule

Method
& \multicolumn{3}{c|}{PCTSP2K}
& \multicolumn{3}{c|}{PCTSP5K}
& \multicolumn{3}{c|}{PCTSP10K}
& \multicolumn{3}{c}{PCTSP20K}
\\

& Obj. $\downarrow$ & Gap & Time
& Obj. $\downarrow$ & Gap & Time
& Obj. $\downarrow$ & Gap & Time
& Obj. $\downarrow$ & Gap & Time
\\
\midrule

LKH-3 (Reference)
& 27.28 & - & -
& 44.39 & - & -
& 66.65 & - & -
& 94.99 & - & - \\

GLOP-G
& \multicolumn{3}{c|}{OOT}
& 49.55 & 11.63 & 1.5\,s
& \multicolumn{3}{c|}{OOT}
& \multicolumn{3}{c}{OOT} \\

GLOP-S
& \multicolumn{3}{c|}{OOT}
& 49.39 & 11.26 & 6.9\,s
& \multicolumn{3}{c|}{OOT}
& \multicolumn{3}{c}{OOT} \\

Attention Model
& 58.55 & 114.62 & 1.6\,s
& 109.38 & 146.41 & 4.0\,s
& 174.96 & 162.51 & 7.4\,s
& \multicolumn{3}{c}{OOM} \\

DeepACO
& 40.68 & 49.12 & 1.3\,min
& 78.98 & 77.92 & 3.4\,min
& \multicolumn{3}{c|}{OOM}
& \multicolumn{3}{c}{OOM} \\

OR-Tools PCTSP
& 29.98 & \textbf{9.88} & 7.5\,min
& 59.64 & 34.35 & 7.5\,min
& 85.31 & 27.98 & 7.5\,min
& 125.02 & 31.62 & 12.8\,min \\

\rowcolor{gray!35}
JI+Refine (Ours)
& 30.99 & 13.62 & \textbf{0.5\,s}
& 48.37 & \textbf{9.01} & \textbf{1.3\,s}
& 70.02 & \textbf{5.05} & \textbf{2.9\,s}
& 101.30 & \textbf{6.65} & \textbf{6.4\,s} \\

\bottomrule

\end{tabular}
}
\end{table}

\subsection{Effectiveness as a Solver Initialization}
\label{sec:solver_initialization}

We evaluate whether JI can improve downstream optimization by providing more effective initial solutions.
Specifically, we compare each solver in its original setting with the corresponding solver augmented by JI initialization.
All reported runtimes include the complete JI pipeline.
Table~\ref{tab:solver_init} reports the final relative gaps and total solving times.

\begin{table}[ht]
\centering
\caption{Search performance with different initializations and refinement procedures. Bold values indicate the better result within each solver pair before rounding.}
\label{tab:solver_init}
\vspace{2pt}
\footnotesize
\setlength{\tabcolsep}{3pt}
\renewcommand{\arraystretch}{1.05}

\begin{tabular}{lrrrrrrrrrrrr}
\toprule
& \multicolumn{2}{c}{1K}
& \multicolumn{2}{c}{5K}
& \multicolumn{2}{c}{10K}
& \multicolumn{2}{c}{20K}
& \multicolumn{2}{c}{50K}
& \multicolumn{2}{c}{100K} \\
\cmidrule(lr){2-3}
\cmidrule(lr){4-5}
\cmidrule(lr){6-7}
\cmidrule(lr){8-9}
\cmidrule(lr){10-11}
\cmidrule(lr){12-13}
Solver
& Gap & Time
& Gap & Time
& Gap & Time
& Gap & Time
& Gap & Time
& Gap & Time \\
\midrule

2-opt
& 11.22 & 0.03\,s
& 11.74 & 0.3\,s
& 11.70 & 0.5\,s
& 11.64 & 1.5\,s
& 11.70 & 9.9\,s
& 11.67 & 42.9\,s \\

JI+2-opt
& \textbf{8.63} & \textbf{0.02\,s}
& \textbf{9.24} & \textbf{0.2\,s}
& \textbf{9.24} & \textbf{0.1\,s}
& \textbf{9.21} & \textbf{0.2\,s}
& \textbf{9.21} & \textbf{0.8\,s}
& \textbf{9.10} & \textbf{3.3\,s} \\
\midrule

VND
& 5.22 & 4.3\,s
& 5.57 & 35.5\,s
& 5.51 & 42.5\,s
& 5.78 & 42.4\,s
& 6.64 & 1.4\,min
& 8.17 & 1.2\,min \\

JI+VND
& \textbf{4.44} & \textbf{2.7\,s}
& \textbf{4.77} & \textbf{24.3\,s}
& \textbf{4.85} & \textbf{28.9\,s}
& \textbf{5.05} & \textbf{35.4\,s}
& \textbf{5.93} & \textbf{34.9\,s}
& \textbf{6.08} & \textbf{1.1\,min} \\
\midrule

ILS
& 6.99 & 1.4\,s
& 8.76 & 41.3\,s
& 9.18 & 40.1\,s
& 10.03 & 41.8\,s
& 10.74 & 1.4\,min
& 11.21 & 1.4\,min \\

JI+ILS
& \textbf{6.59} & \textbf{0.9\,s}
& \textbf{8.49} & \textbf{16.9\,s}
& \textbf{8.75} & \textbf{12.9\,s}
& \textbf{8.90} & \textbf{31.8\,s}
& \textbf{9.04} & \textbf{1.3\,min}
& \textbf{9.04} & \textbf{1.3\,min} \\
\midrule

GLS
& 10.38 & 2.8\,s
& 11.22 & 43.7\,s
& 11.44 & 43.4\,s
& 11.53 & 40.3\,s
& 11.68 & 45\,s
& 11.66 & 1.2\,min \\

JI+GLS
& \textbf{8.12} & \textbf{1.5\,s}
& \textbf{9.00} & \textbf{14.3\,s}
& \textbf{9.07} & \textbf{16.6\,s}
& \textbf{9.13} & \textbf{34.6\,s}
& \textbf{9.19} & \textbf{35.3\,s}
& \textbf{9.10} & \textbf{54.8\,s} \\
\midrule

\rowcolor{gray!35}
JI+Refine
& \textbf{2.56} & \textbf{0.04\,s}
& \textbf{3.48} & \textbf{0.2\,s}
& \textbf{3.51} & \textbf{0.5\,s}
& \textbf{3.68} & \textbf{1.4\,s}
& \textbf{3.74} & \textbf{4.8\,s}
& \textbf{3.80} & \textbf{12.2\,s} \\
\bottomrule
\end{tabular}
\vspace{-5pt}
\end{table}

The results show that JI consistently improves both solution quality and computational efficiency across all refinement procedures and problem scales.
Compared with the original solvers, JI-enhanced solvers achieve lower final gaps with shorter overall runtimes, demonstrating that JI provides effective initializations for downstream optimization. This verifies that improving initialization can reduce the burden of subsequent optimization. 

We further evaluate our dedicated refinement strategy under the same JI-generated initialization.
Compared with existing refinement procedures using identical initializations, the proposed refinement achieves better solution quality, showing that effective initialization and specialized refinement provide complementary benefits.

Additional initialization analyses, including matched-gap and greedy-start comparisons as well as search-budget evaluations, are provided in Appendix~\ref{sec:appendix_matched_gap}--\ref{sec:appendix_greedy_starts}.

\subsection{Ablation Study}
\label{sec:ablation}

We perform ablation studies on uniform TSP instances by removing each component of JI individually, including compression, compressed-space solving, and recovery. The gap increase relative to the full model is reported in parentheses. As shown in Table~\ref{tab:path_ablation}, removing compression or recovery consistently increases the final gap across different scales, with over 1\% degradation in large-scale instances, while removing the solving stage causes smaller changes. These results verify that compression and recovery are essential for generating effective initializations, and the complete JI pipeline enables optimization to start from more favorable regions of the solution space.

\begin{table}[ht]

\centering
\caption{JI ablation on uniform TSP instances. Numbers in parentheses denote the gap increase relative to the full model. All results are averaged over three seeds.}
\label{tab:path_ablation}
\vspace{2pt}
\small
\setlength{\tabcolsep}{3.5pt}
\renewcommand{\arraystretch}{1.05}

\begin{tabular}{lrrrrrrrr}
\toprule
& \multicolumn{2}{c}{Full} 
& \multicolumn{2}{c}{w/o compress}
& \multicolumn{2}{c}{w/o solve}
& \multicolumn{2}{c}{w/o restore}\\
\cmidrule(lr){2-3}
\cmidrule(lr){4-5}
\cmidrule(lr){6-7}
\cmidrule(l){8-9}

Scale 
& Gap (\%) & Time (s)
& Gap (\%) & Time (s)
& Gap (\%) & Time (s)
& Gap (\%) & Time (s)\\
\midrule

1K   
& 2.84 & 0.03  
& 3.59 (+0.75) & 0.04  
& 3.39 (+0.55) & 0.05  
& 3.90 (+1.06) & 0.03 \\

5K   
& 3.62 & 0.16  
& 4.26 (+0.64) & 0.18  
& 3.68 (+0.06) & 0.16  
& 4.44 (+0.82) & 0.17 \\

10K  
& 3.49 & 0.36  
& 4.83 (+1.34) & 0.47  
& 3.60 (+0.11) & 0.35  
& 4.66 (+1.17) & 0.41 \\

20K  
& 3.70 & 0.89  
& 4.88 (+1.18) & 1.20  
& 3.74 (+0.04) & 0.91  
& 4.87 (+1.17) & 1.11 \\

50K  
& 3.67 & 4.23 
& 4.82 (+1.15) & 7.23 
& 4.25 (+0.58) & 4.12 
& 4.96 (+1.29) & 5.30 \\
100K 
& 3.78 & 12.56 
& 5.00 (+1.22) & 24.91 
& 4.34 (+0.56) & 12.75
& 5.08 (+1.30) & 16.89 \\

\bottomrule
\end{tabular}
\vspace{-5pt}
\end{table}

\section{Conclusion}
In this work, we revisit large-scale routing optimization from the perspective of where optimization begins. 
Instead of continuously increasing the complexity of optimization procedures, we show that a more effective starting point can fundamentally reduce the difficulty of downstream search.  Based on this insight, we propose \textbf{Just Initialize}, a training-free and solver-independent initialization component that separates solution discovery from solution optimization. By exploring coarse global structures in a compact space and recovering them as optimization-friendly starting points, Just Initialize enables existing solvers to focus on refining promising regions rather than searching the entire solution space.

Extensive experiments on TSP, CVRP, VRPTW, and PCTSP demonstrate that Just Initialize scales to instances from 1K to 100K nodes while maintaining competitive solution quality with substantially reduced computational cost. 
These results reveal that initialization can serve as an effective mechanism for navigating large combinatorial search spaces, providing a complementary direction to improving optimization algorithms themselves.

\section{Acknowledgments}
This paper is supported by National Key Research and Development Program of China (Grant No. 2024YFB3311900), Beijing Natural Science Foundation (Grant No. L241018) and Beijing Nova Program (Grant No. 2024085).
\newpage
\subsection*{AI use statement}

In this work, we used generative AI tools to aid or polish writing, retrieve and discover related work, and draft sections of the paper.

We have not used generative AI tools for the following other required-disclosure tasks that were relevant to our research but performed entirely by the authors: helping develop theoretical models or conceptual frameworks, proposing or refining hypotheses, and designing or providing feedback on research methodology or experiments.

The remaining required-disclosure tasks are not applicable to this work because our study did not involve them: assisting in the writing of proofs, providing critical ingredients for proving mathematical claims, and formulating mathematical claims.

Additionally, we used generative AI tools for the following recommended-disclosure tasks: drafting parts of a research paper, editing a research paper to improve readability, formatting references, and creating or editing software code (for utility scripts and visualisation functions).

We have reviewed all AI-assisted work. Specifically:
\begin{itemize}
    \item All AI-generated code was tested on multiple runs, verified for correctness against expected outputs, and peer-reviewed by at least two authors.
    \item The cleaned dataset was manually inspected for consistency and missing-value handling after AI-assisted preprocessing.
    \item AI-drafted text was substantially revised to match our scientific voice and checked for factual accuracy against our experimental results.
\end{itemize}

We take responsibility for the final content of this work, including text, claims, or artifacts produced with the aid of generative AI. The authors alone developed the core research ideas, designed the experiments, and interpreted the final results, without AI involvement.

\subsection*{Ethics statement}

This paper studies training-free initialization methods for large-scale routing optimization. It does not involve human subjects, personal data, or privacy-sensitive information. All datasets used are publicly available and contain mathematical problems. We foresee no ethical concerns and declare no conflicts of interest.

\subsection*{Reproducibility statement}

We describe Just Initialize in Section~\ref{sec:methodology},
with algorithmic pseudocode and parameter settings in
Appendix~\ref{app:implementation}.
Dataset construction procedures, generation seeds, and
additional solver settings are provided in
Appendix~\ref{app:datasets_parameters}.
Experimental hardware, evaluation metrics, and runtime
measurement conventions are described in
Section~\ref{sec:experiment_settings}.

\bibliography{iclr2027_conference}
\bibliographystyle{iclr2027_conference}

\appendix
\section{Baselines and Benchmarks}

\subsection{Baselines}
We compare Just Initialize with the methods in Tables~\ref{tab:tsp_cvrp_main} and~\ref{tab:large_scale_variants}. Each description below summarizes a method's main strength and the corresponding computational or modeling tradeoff.

\paragraph{Classical solvers and metaheuristics.}
LKH-3~\citep{helsgaun2017lkh3} combines Lin--Kernighan search with problem-specific extensions and repeated local search.
Concorde~\citep{concorde} provides exact TSP solutions and optimality certificates through branch-and-cut, at a computational cost that rises sharply with instance size.
HGS~\citep{vidal2022hgs} combines population diversity with specialized CVRP local search, while crossover and route improvement add search overhead.
OR-Tools and its PCTSP configuration~\citep{ortools} offer flexible routing constraints and established search operators, with solution quality depending on the chosen search strategy and time limit.
VROOM~\citep{vroom} rapidly constructs and improves feasible vehicle routes with capacities and time windows, trading exact guarantees for heuristic search speed.
ALNS~\citep{ropke2006alns} explores route changes through adaptive removal and repair operators with repeated reconstruction.
PCTSP-ILS iteratively changes the visited-customer set and tour to explore prize--penalty tradeoffs, while repeated local improvement increases runtime.

\paragraph{Neural construction and diffusion methods.}
POMO~\citep{kwon2020pomo} exploits multiple equivalent starting points to produce diverse candidate routes, while eightfold augmentation multiplies inference work relative to the unaugmented configuration.
LEHD~\citep{luo2023neural} uses a heavy decoder to make context-aware construction decisions on large instances, while sequential decoding raises inference cost.
INViT~\citep{fang2024invit} uses invariant nested views to transfer routing decisions across scales and distributions, while processing multiple views adds computation.
SIL~\citep{luo2024sil} learns from solutions improved by local reconstruction and uses linear-complexity attention for scale, while generating and retraining on pseudo-labels adds training effort.
L2C-Insert~\mbox{\citep{luo2025insert}} can place a customer at any valid position in a partial route, while evaluating insertion positions costs more than simple append decisions.
The Attention Model~\citep{kool2019attention} learns route construction directly with attention, while autoregressive decoding remains sequential as the instance grows.
DIFUSCO~\citep{sun2023difusco} uses diffusion-generated edge scores to guide global tour construction, while denoising and tour recovery add inference stages.

\paragraph{Decomposition and learned search.}
GLOP~\citep{ye2024glop} combines global partitioning with local neural construction to scale to large routing instances, while the final tour depends on the quality of the partition.
GLOP-G~\citep{ye2024glop} uses greedy partition decoding to reduce sampling work, while a single partition offers less exploration than sampled decoding.
UDC~\citep{zheng2024udc} jointly learns division and solution of subproblems across routing tasks, while reunion must reconcile decisions made in separate parts.
H-TSP~\citep{pan2023htsp} constructs large tours through hierarchical node selection and local routing, while early subset decisions influence later connections.
L2D~\citep{li2021learning} learns which subroutes to delegate to a local solver for focused improvement, while repeated delegated searches add runtime.
TTPL~\citep{chen2025ttpl} adapts a learned policy to test-instance features to address distribution shift, while the projection step adds test-time work.
DeepACO~\citep{ye2023deepaco} combines learned heuristic information with ant-colony exploration, while more ants and iterations increase the search budget.

\subsection{Benchmarks}
Let $N=\{1,\ldots,n\}$ be the set of cities or customers. For a node set $V$, let $A(V)=\{(i,j):i,j\in V,\ i\ne j\}$ be its directed arc set and let $d_{ij}$ be the instance-defined travel distance. The formulations below describe the feasible routes and objectives; Appendix~D specifies how their attributes and reference values are generated.

For synthetic instances, coordinates and distance-related quantities use a $10^6$ integer export; VRPTW time attributes and PCTSP prizes, penalties, and quota are scaled consistently, while demands and capacities retain their original units. Reported synthetic objectives are divided by $10^6$. Public TSP instances use their source distance rules and unscaled objectives. For an instance objective $C$ and its recorded reference $C_{\mathrm{ref}}$, the reported gap is $100(C-C_{\mathrm{ref}})/C_{\mathrm{ref}}$ percent; dataset composition and reference procedures are detailed in Appendix~D.

\noindent\textit{Definition A.1 (TSP).}
For the traveling salesman problem, set $V=N$ and let $x_{ij}=1$ when the tour traverses arc $(i,j)$. Every city has one predecessor and one successor, and the subtour constraints connect them into a single tour:
\begin{align}
\min_x\quad & \sum_{(i,j)\in A(V)} d_{ij}x_{ij}
    \label{eq:app_tsp_obj}\\
\text{s.t.}\quad
& \sum_{j\in V\setminus\{i\}}x_{ij}=1,
    && \forall i\in V, \label{eq:app_tsp_out}\\
& \sum_{j\in V\setminus\{i\}}x_{ji}=1,
    && \forall i\in V, \label{eq:app_tsp_in}\\
& \sum_{\substack{i,j\in S\\i\ne j}}x_{ij}\le |S|-1,
    && \forall S\subset V,\ 2\le |S|\le n-1, \label{eq:app_tsp_subtour}\\
& x_{ij}\in\{0,1\},
    && \forall (i,j)\in A(V). \label{eq:app_tsp_domain}
\end{align}
The synthetic TSP benchmarks contain 96 uniform and 144 extended-distribution instances; the additional 42 public instances retain their original distance definitions.

\noindent\textit{Definition A.2 (CVRP).}
For the capacitated vehicle routing problem, let $V=\{0\}\cup N$, where $0$ is the depot, $q_i>0$ is customer demand, and $Q$ is vehicle capacity. The binary variable $x_{ij}$ selects arc $(i,j)$, and the continuous variable $u_i$ is the load delivered up to customer $i$ on its route. The number of depot-to-depot routes is chosen by the solution:
\begin{align}
\min_{x,u}\quad & \sum_{(i,j)\in A(V)}d_{ij}x_{ij}
    \label{eq:app_cvrp_obj}\\
\text{s.t.}\quad
& \sum_{j\in V\setminus\{i\}}x_{ij}=1,
    && \forall i\in N, \label{eq:app_cvrp_out}\\
& \sum_{j\in V\setminus\{i\}}x_{ji}=1,
    && \forall i\in N, \label{eq:app_cvrp_in}\\
& \sum_{j\in N}x_{0j}=\sum_{i\in N}x_{i0},
    && \label{eq:app_cvrp_depot}\\
& u_j\ge u_i+q_j-Q(1-x_{ij}),
    && \forall i,j\in N,\ i\ne j, \label{eq:app_cvrp_load}\\
& q_i\le u_i\le Q,
    && \forall i\in N, \label{eq:app_cvrp_capacity}\\
& x_{ij}\in\{0,1\},
    && \forall (i,j)\in A(V). \label{eq:app_cvrp_domain}
\end{align}
Positive demands and the load constraints also rule out customer-only cycles. The main CVRP comparison uses 80 instances at five scales from 5K to 100K customers, each with one additional depot.

\noindent\textit{Definition A.3 (VRPTW).}
The vehicle routing problem with time windows retains customer demands $q_i$ and capacity $Q$, and adds a service duration $s_i$ and a service-start window $[a_i,b_i]$ at each customer. To represent route departures and returns, use a start depot $0$ and an end-depot copy $n+1$. The arc set $A^{\mathrm{tw}}$ contains arcs from $0$ to customers, between distinct customers, and from customers to $n+1$.
The end copy has the depot's location and window, so $d_{i,n+1}=d_{i0}$ and $\tau_{i,n+1}=\tau_{i0}$. Here $t_i$ is the continuous service start, $\tau_{ij}$ is travel time, $u_i$ is continuous cumulative load, and $M_{ij}$ is a valid time-propagation bound. Waiting is allowed before service begins:
\begin{align}
\min_{x,u,t}\quad & \sum_{(i,j)\in A^{\mathrm{tw}}}d_{ij}x_{ij}
    \label{eq:app_vrptw_obj}\\
\text{s.t.}\quad
& \sum_{j:(i,j)\in A^{\mathrm{tw}}}x_{ij}=1,
    && \forall i\in N, \label{eq:app_vrptw_out}\\
& \sum_{j:(j,i)\in A^{\mathrm{tw}}}x_{ji}=1,
    && \forall i\in N, \label{eq:app_vrptw_in}\\
& \sum_{j\in N}x_{0j}=\sum_{i\in N}x_{i,n+1},
    && \label{eq:app_vrptw_depot}\\
& u_j\ge u_i+q_j-Q(1-x_{ij}),
    && \forall i,j\in N,\ i\ne j, \label{eq:app_vrptw_load}\\
& q_i\le u_i\le Q,
    && \forall i\in N, \label{eq:app_vrptw_capacity}\\
& t_j\ge t_i+s_i+\tau_{ij}-M_{ij}(1-x_{ij}),
    && \forall (i,j)\in A^{\mathrm{tw}}, \label{eq:app_vrptw_time}\\
& a_i\le t_i\le b_i,
    && \forall i\in N, \label{eq:app_vrptw_windows}\\
& t_0=a_0,\quad a_0\le t_{n+1}\le b_0,
    && \label{eq:app_vrptw_depot_window}\\
& x_{ij}\in\{0,1\},
    && \forall (i,j)\in A^{\mathrm{tw}}. \label{eq:app_vrptw_domain}
\end{align}
We take $s_0=0$, $a_{n+1}=a_0$, and $M_{ij}=\max\{0,b_i+s_i+\tau_{ij}-a_j\}$; the benchmark has 72 instances from 1K to 20K customers, each with one physical depot.

\noindent\textit{Definition A.4 (PCTSP).}
For the prize-collecting traveling salesman problem, let $V=\{0\}\cup N$, with depot $0$, prize $p_i$, omission penalty $\lambda_i$, and required collected prize $P$. The binary variable $y_i$ records whether customer $i$ is visited, while $x_{ij}$ selects tour arcs. A single depot tour minimizes travel plus penalties for omitted customers:
\begin{align}
\min_{x,y}\quad
& \sum_{(i,j)\in A(V)}d_{ij}x_{ij}
  +\sum_{i\in N}\lambda_i(1-y_i)
    \label{eq:app_pctsp_obj}\\
\text{s.t.}\quad
& \sum_{j\in V\setminus\{i\}}x_{ij}=y_i,
    && \forall i\in N, \label{eq:app_pctsp_out}\\
& \sum_{j\in V\setminus\{i\}}x_{ji}=y_i,
    && \forall i\in N, \label{eq:app_pctsp_in}\\
& \sum_{j\in N}x_{0j}=\sum_{i\in N}x_{i0}=1,
    && \label{eq:app_pctsp_depot}\\
& \sum_{i\in S}\sum_{j\in V\setminus S}x_{ij}\ge y_k,
    && \forall \emptyset\ne S\subseteq N,\ k\in S,
       \label{eq:app_pctsp_connect}\\
& \sum_{i\in N}p_i y_i\ge P,
    && \label{eq:app_pctsp_prize}\\
& x_{ij}\in\{0,1\},
    && \forall (i,j)\in A(V), \label{eq:app_pctsp_arc_domain}\\
& y_i\in\{0,1\},
    && \forall i\in N. \label{eq:app_pctsp_visit_domain}
\end{align}
The PCTSP benchmark contains 72 instances from 1K to 20K optional customers, with $P=1$ before integer export.

\section{Implementation Details of Just Initialize}
\label{app:implementation}

\subsection{Algorithmic Implementation}
\label{app:implementation_algorithm}

Algorithm~\ref{alg:ji_impl} summarizes the implementation of Just Initialize.
The procedure constructs an initial solution in the compact space, improves it
under a restricted geometric neighborhood, recovers a structurally compatible
solution in the original space, and performs a bounded local refinement.
During compression, each compact unit additionally retains an ordered sequence
of its original nodes, which is used to instantiate the unit during recovery.

\begin{algorithm}[ht]
\caption{Implementation of Just Initialize}
\label{alg:ji_impl}
\KwIn{Instance $I$; compression scale $b$; candidate width $K_c$;
search budget $T_c$; evaluation samples $S$; budgets $B,B_\mu$;
temperature $\tau$}
\KwOut{Refined initialization $\pi_{\mathrm{JI}}$}

$(\tilde I,M,\Phi_s)\leftarrow\textsc{Compress}(I,b)$ using
Algorithm~\ref{alg:compression}\;
$\mathcal Q\leftarrow\textsc{BuildSpatialIndex}(\{z_i:C_i\in\tilde I\})$\;

$\tilde\pi\leftarrow\emptyset$,\quad
$\mathcal U\leftarrow\textsc{RequiredCompactUnits}(\tilde I)$\;
\While{$\mathcal U\neq\emptyset$}{
    $\mathcal C\leftarrow
    \textsc{QueryCandidates}(\mathcal Q,\tilde\pi,\mathcal U,K_c)$\;
    $\mathcal A\leftarrow
    \{(C_i,t):C_i\in\mathcal C,\,
    t\in\mathcal T_i,\,
    \textsc{FeasibleAppend}(\tilde\pi,C_i,t,\tilde K)\}$\;
    \eIf{$\mathcal A=\emptyset$}{
        $\tilde\pi\leftarrow\textsc{StartNewRoute}(\tilde\pi)$\;
    }{
        $(C^\star,t^\star)\leftarrow
        \arg\min_{(C_i,t)\in\mathcal A}
        \textsc{ConnectionCost}(\tilde\pi,C_i,t,\tilde C)$\;
        $\tilde\pi\leftarrow\tilde\pi\oplus(C^\star,t^\star)$,\quad
        $\mathcal U\leftarrow\mathcal U\setminus\{C^\star\}$\;
    }
}

\For{$k\leftarrow1$ \KwTo $T_c$}{
    $C_i\leftarrow\textsc{SelectCompactUnit}(\tilde\pi)$,\quad
    $\mathcal C_i\leftarrow\textsc{QueryCandidates}(\mathcal Q,C_i,K_c)$\;
    $\widetilde{\mathcal N}\leftarrow
    \textsc{GenerateCompactNeighbors}(\tilde\pi,C_i,\mathcal C_i)
    \cap\Omega(\tilde I)$\;

    \ForEach{$\tilde\pi'\in\widetilde{\mathcal N}\cup\{\tilde\pi\}$}{
        $\widehat J_B(\tilde\pi')\leftarrow0$\;
        \For{$s\leftarrow1$ \KwTo $S$}{
            $\mathcal P_s\leftarrow
            \textsc{RecoveryCandidates}(\tilde\pi',M)$\;
            $\pi_0^{(s)}\sim
            \dfrac{\exp(g(\pi;\tilde\pi',M)/\tau)}
            {\sum_{\bar\pi\in\mathcal P_s}
            \exp(g(\bar\pi;\tilde\pi',M)/\tau)}$\;
            $\pi^{(s)}\leftarrow\pi_0^{(s)}$\;
            \For{$r\leftarrow1$ \KwTo $B$}{
                $\mathcal N\leftarrow
                \textsc{CandidateRestrict}(\mathcal N_I(\pi^{(s)}))
                \cap\Omega(I)$\;
                $\pi'\leftarrow
                \arg\min_{\bar\pi\in\mathcal N}C_I(\bar\pi)$\;
                \If{$\mathcal N=\emptyset$ \textbf{ or }
                    $C_I(\pi')\geq C_I(\pi^{(s)})$}{\textbf{break}}
                $\pi^{(s)}\leftarrow\pi'$\;
            }
            $\widehat J_B(\tilde\pi')\leftarrow
            \widehat J_B(\tilde\pi')+C_I(\pi^{(s)})/S$\;
        }
    }
    $\tilde\pi\leftarrow
    \arg\min_{\bar\pi\in\widetilde{\mathcal N}\cup\{\tilde\pi\}}
    \widehat J_B(\bar\pi)$\;
}

$\mathcal P\leftarrow\textsc{RecoveryCandidates}(\tilde\pi,M)$,\quad
$\pi_0^\star\sim
\dfrac{\exp(g(\pi;\tilde\pi,M)/\tau)}
{\sum_{\bar\pi\in\mathcal P}\exp(g(\bar\pi;\tilde\pi,M)/\tau)}$\;

$\pi\leftarrow\pi_0^\star$\;
\For{$r\leftarrow1$ \KwTo $B_\mu$}{
    $\mathcal N_\mu\leftarrow
    \textsc{CandidateRestrict}(\mathcal N_I(\pi))\cap\Omega(I)$\;
    $\pi'\leftarrow\arg\min_{\bar\pi\in\mathcal N_\mu}C_I(\bar\pi)$\;
    \If{$\mathcal N_\mu=\emptyset$ \textbf{ or } $C_I(\pi')\geq C_I(\pi)$}
        {\textbf{break}}
    $\pi\leftarrow\pi'$\;
}
\Return $\pi$\;
\end{algorithm}

The compact representation follows the same structure for all four problem classes, while the attributes retained by a unit reflect the corresponding constraints. TSP primarily uses the local node sequence and its boundary geometry. CVRP additionally retains the aggregated demand of each unit, PCTSP retains aggregated prize and penalty information, and VRPTW augments the unit representation with demand and temporal boundary information. Accordingly, the feasibility test in Algorithm~\ref{alg:ji_impl} checks only the resources relevant to the current problem. 

The compact objective $F_P$ is likewise instantiated according to the routing problem. For TSP, it measures the closed-tour cost induced by the ordered compact fragments. For CVRP, it is the sum of depot-to-depot route costs, with capacity-infeasible routes excluded. VRPTW uses the same distance-based route objective while additionally enforcing capacity and time-window feasibility during concatenation. For PCTSP, the travel cost of the selected compact tour is combined with the penalties of unselected units, and plans that do not satisfy the prize quota are infeasible. The incremental score $\Delta_P$ used during construction follows the same information: required-node problems are primarily guided by incremental connection cost, while optional unit selection also accounts for the prize--penalty structure.

Compact-space search uses the same restricted geometric neighborhood in all cases. Exchange, relocation, and reversal modify the arrangement of compact units, while optional-node instances additionally allow the selected set to change. These variations affect the candidate set but not the subsequent search procedure.

Downstream evaluation is applied only to a small shortlist rather than to every compact neighbor. For deterministic refinement, this evaluation corresponds to a single trajectory of $\widehat J_{B,P}$. Recovery itself expands each compact route by concatenating the stored local sequences according to their selected orientations. The resulting original-space solution is then refined under the same problem-dependent objective and feasibility conditions until no improving candidate remains or the budget $B_\mu$ is exhausted.
\subsection{Structural Consistency Score for Recovery}
\label{app:structural_score}

We provide the explicit formulation of the structural consistency
score $g(\pi;\tilde{\pi},M)$ used during initialization recovery.
The purpose of the score is to map the geometric information retained
by the compact representation to a scalar measure of compatibility
for candidate recovered solutions.

Let a compact solution be represented as
\begin{equation}
\tilde{\pi}
=
\bigl(
(C_{i_1},t_{i_1}),
\ldots,
(C_{i_m},t_{i_m})
\bigr),
\end{equation}
where $C_i$ denotes a compact unit and $t_i\in T_i$ is its selected
boundary state. For each unit, the correspondence information $M$
stores its original nodes, geometric summary, boundary states, and
local geometric information. We denote by $q_i(t_i)$ the ordered
local node sequence associated with boundary state $t_i$.

For a boundary state $t_i$, let
$a_i^{\mathrm{in}}(t_i)$ and $a_i^{\mathrm{out}}(t_i)$ denote the
reference entry and exit locations of compact unit $C_i$,
respectively. For a recovered solution $\pi$, let
$u_i^{\mathrm{in}}(\pi)$ and $u_i^{\mathrm{out}}(\pi)$ denote the
actual entry and exit nodes used by $\pi$.
Furthermore, let
$\tilde{c}_{ij}(t_i,t_j)$ denote the connection cost between
$C_i$ and $C_j$ estimated from their compact representations.

\paragraph{Geometry-guided boundary selection.}
For two consecutive compact units $C_i$ and $C_j$, consider a
candidate exit node $u\in C_i$ and entry node $v\in C_j$.
We define their normalized geometric discrepancy as
\begin{equation}
\left\{
\begin{aligned}
\delta_{\mathrm{out}}(u;i)
&=
\frac{
d\!\left(u,a_i^{\mathrm{out}}(t_i)\right)
}{
\operatorname{diam}(C_i)+\varepsilon
},
\\[4pt]
\delta_{\mathrm{in}}(v;j)
&=
\frac{
d\!\left(v,a_j^{\mathrm{in}}(t_j)\right)
}{
\operatorname{diam}(C_j)+\varepsilon
},
\\[4pt]
\delta_{\mathrm{conn}}(u,v;i,j)
&=
\frac{
\left|
d(u,v)-\tilde{c}_{ij}(t_i,t_j)
\right|
}{
\tilde{c}_{ij}(t_i,t_j)+\varepsilon
},
\\[4pt]
\ell_{\mathrm{geo}}(u,v;i,j)
&=
\frac{1}{3}
\left[
\delta_{\mathrm{out}}(u;i)
+
\delta_{\mathrm{in}}(v;j)
+
\delta_{\mathrm{conn}}(u,v;i,j)
\right].
\end{aligned}
\right.
\label{eq:geo_components}
\end{equation}
where $d(\cdot,\cdot)$ is the distance defined by the original
routing instance, $\operatorname{diam}(C_i)$ is the geometric
diameter of compact unit $C_i$, and $\varepsilon>0$ is a small
constant used only for numerical stability.

The three terms measure, respectively, the deviation of the selected
exit from the reference exit, the deviation of the selected entry
from the reference entry, and the mismatch between the realized
inter-unit connection and its compact-space estimate. All terms are
normalized to make their scales comparable.

Accordingly, the geometrically preferred connection between two
consecutive compact units is
\begin{equation}
\left(
u_{ij}^{*},v_{ij}^{*}
\right)
=
\arg\min_{
u\in C_i,\,
v\in C_j
}
\ell_{\mathrm{geo}}(u,v;i,j).
\label{eq:boundary_selection}
\end{equation}
In practice, the lowest-discrepancy pairs are used to construct the
recovery candidate set rather than exhaustively enumerating all
possible node pairs.

\paragraph{Inter-unit geometric consistency.}
Let $\mathcal{E}_{\tilde{\pi}}$ denote the set of consecutive
compact-unit pairs induced by $\tilde{\pi}$, taken route-wise for
multi-route problems. For a complete recovered solution $\pi$, we
define
\begin{equation}
L_{\mathrm{geo}}
(\pi;\tilde{\pi},M)
=
\frac{1}{
|\mathcal{E}_{\tilde{\pi}}|
}
\sum_{(i,j)\in\mathcal{E}_{\tilde{\pi}}}
\ell_{\mathrm{geo}}
\left(
u_i^{\mathrm{out}}(\pi),
u_j^{\mathrm{in}}(\pi);
i,j
\right).
\label{eq:global_geo_loss}
\end{equation}
A smaller $L_{\mathrm{geo}}$ indicates that the recovered solution
better preserves the boundary geometry and inter-unit connections
represented in the compact solution.

\paragraph{Intra-unit sequence consistency.}
The compact representation also retains an ordered local sequence
for each unit. Let $\pi_i$ denote the subsequence of original nodes
belonging to $C_i$ in recovered solution $\pi$, and let $E(q)$ denote
the set of consecutive node pairs in sequence $q$. We define
\begin{equation}
L_{\mathrm{seq}}
(\pi;\tilde{\pi},M)
=
\frac{1}{
|\mathcal{C}_{\tilde{\pi}}|
}
\sum_{C_i\in\mathcal{C}_{\tilde{\pi}}}
\left(
1-
\frac{
\left|
E(\pi_i)
\cap
E(q_i(t_i))
\right|
}{
\max(1,|C_i|-1)
}
\right),
\label{eq:sequence_loss}
\end{equation}
where $\mathcal{C}_{\tilde{\pi}}$ denotes the set of compact units
selected by $\tilde{\pi}$. Thus,
$L_{\mathrm{seq}}=0$ when all stored local adjacencies are preserved,
and the discrepancy increases as the recovered solution departs from
the local structure retained during compression.

\paragraph{Structural consistency score.}
Since recovery candidates are restricted to
$\Phi_s^{-1}(\tilde{\pi};M)$, the ordering of compact units specified
by $\tilde{\pi}$ is preserved by construction. The remaining
compatibility therefore concerns how the compact structure is
instantiated at the original-node level. We define
\begin{equation}
g(\pi;\tilde{\pi},M)
=
-\frac{1}{2}
\left(
L_{\mathrm{geo}}
(\pi;\tilde{\pi},M)
+
L_{\mathrm{seq}}
(\pi;\tilde{\pi},M)
\right).
\label{eq:structural_consistency}
\end{equation}

A larger value of $g$ therefore corresponds to a recovered solution
that more faithfully preserves the geometric connections, boundary
configurations, and local node structure represented by the compact
solution. Because all discrepancy terms are normalized, equal
weighting is used without introducing additional problem-specific
hyperparameters.

The resulting score is used in the recovery distribution
\begin{equation}
\Pr(
\pi_0^{*}=\pi
\mid
\tilde{\pi}^{*},M
)
=
\frac{
\exp
\left(
g(\pi;\tilde{\pi}^{*},M)/\tau
\right)
}{
\displaystyle
\sum_{\bar{\pi}
\in
\Phi_s^{-1}(\tilde{\pi}^{*};M)}
\exp
\left(
g(\bar{\pi};\tilde{\pi}^{*},M)/\tau
\right)
},
\end{equation}
so that geometrically more consistent candidates receive higher
recovery probability while the temperature $\tau$ controls the
degree of concentration.
\subsection{Parameter Settings}
\label{app:implementation_parameters}

\begin{table}[ht]
\centering
\caption{Shared implementation parameters of Just Initialize.
$m$ denotes the number of compact units.}
\label{tab:ji_shared_parameters}

\footnotesize
\setlength{\tabcolsep}{4pt}
\renewcommand{\arraystretch}{1.12}

\begin{tabular}{@{}
    >{\centering\arraybackslash}m{0.20\linewidth}
    >{\centering\arraybackslash}m{0.25\linewidth}
    m{0.49\linewidth}
@{}}
\toprule

\textbf{Parameter}
& \textbf{Value}
& \centering\arraybackslash\textbf{Role} \\
\midrule

$K_c$ & $\min\{64,m\}$ & Candidate width for compact-solution construction \\
$K_{\mathrm{local}}$ & $\min\{24,m\}$ & Candidate width for compact-space local search \\
$K_{\mathrm{merge}}$ & $\min\{32,m\}$ & Candidate width for route merging, when applicable \\
$S$ & $1$ & Number of refinement trials used to estimate $\widehat J_B$ \\
$\tau$ & Low temperature & Controls the concentration of recovery sampling \\
$B_{\mu}$ & Bounded & Maximum number of post-recovery refinement steps \\
$s$ & $0$ & Random seed for compact-space search \\

\bottomrule
\end{tabular}
\end{table}

Table~\ref{tab:ji_shared_parameters} summarizes the parameters shared across all problem classes. These settings are kept fixed throughout the experiments. The problem-dependent settings primarily involve compression granularity and compact-search budget. For TSP, compact units contain approximately $32$ nodes, while PCTSP uses smaller units of $8$ nodes. CVRP controls the granularity by accumulated demand, with a target load of $0.8Q$ per compact unit. For VRPTW, compression groups up to $40$ local routes, with individual fragments limited to at most $16$ nodes to preserve local temporal structure. The compact-search budget $T_c$ is scaled with the compressed size $m$ and adjusted to the structural complexity of each problem class. All problem-specific settings are fixed within each problem class and are not tuned for individual instances.

\section{Theoretical Analysis}
\label{app:theory}

We analyze Just Initialize under the computational model that
Algorithm~\ref{alg:ji_impl} implements. An instance contains $n$ nodes, which
are grouped into $m$ compact units, so that a unit contains $b=n/m$ nodes on
average; we assume $b>1$. Each unit carries a constant-size summary of the
nodes it contains: its geometry $z_i$, its boundary states $\mathcal T_i$, and,
where the problem requires it, aggregated demand, prize, or temporal
information. Both spaces are searched with candidate-restricted neighborhoods:
a sweep in the original space examines at most $K$ candidate neighbors at each
of the $n$ positions, and a sweep in the compact space examines at most $K_c$
candidate units at each of the $m$ units. The refinement operator is a descent
procedure: it accepts a move only when the objective strictly decreases, and it
terminates when no improving candidate remains or its budget is exhausted. The
analysis below addresses three questions in turn, and each result is stated
with the assumptions it requires and the design choice it supports.

\subsection{Decision Count and Cost of the Compact Stage}
\label{app:theory_efficiency}

\paragraph{Proposition 1 (Decision-space reduction).}
Under candidate-restricted search, one sweep examines at most
\begin{equation}
D_o=nK,\qquad D_c=mK_c
\end{equation}
candidate decisions in the original and compact spaces, respectively. With
$m=n/b$,
\begin{equation}
\frac{D_c}{D_o}=\frac{K_c}{bK},
\label{eq:app_decision_ratio}
\end{equation}
so that when the two candidate widths are of the same order
($K_c=\Theta(K)$), the number of decisions examined per sweep is reduced by the
compression factor $b$.

\paragraph{Proof.}
Each sweep visits every position once and examines at most $K$ (respectively
$K_c$) candidates per position, which gives the two counts in
Eq.~(1). Substituting $m=n/b$ yields Eq.~(\ref{eq:app_decision_ratio}).
\hfill $\square$

\paragraph{Proposition 2 (Cost of a compact decision).}
Let the connection cost of appending a unit to a partial compact plan be
evaluated from the stored summaries of the two units, as in
Algorithm~\ref{alg:ji_impl}. Then this evaluation runs in $O(1)$ time,
independent of the number $b$ of original nodes represented by a unit, and one
construction sweep costs
\begin{equation}
T_{\mathrm{construct}}=O(mK_c),
\end{equation}
whereas evaluating the same number of placements against the full instance
costs $O(nK)$.

\paragraph{Proof.}
The connection cost is a function of the two unit summaries, each of constant
size, hence is evaluated in constant time. Multiplying by the number of
examined candidates gives $O(mK_c)$ for the compact sweep. In the original
space, each candidate evaluation reads both endpoint neighborhoods and thus
costs $O(K)$ per position, giving $O(nK)$ per sweep.
\hfill $\square$

Combining Propositions~1 and~2, the compact stage removes a factor $b$ from
both the number of decisions and, by Eq.~(\ref{eq:app_decision_ratio}), the
geometric work per sweep. The remaining cost of a run decomposes as
\begin{equation}
T_{\mathrm{JI}}
=
T_{\mathrm{compress}}
+O(mK_c)
+\sum_{j=1}^{L}\sum_{s=1}^{S}\Bigl(T_{\mathrm{rec},j,s}
+B\,T_{\mathrm{move}}\Bigr)
+T_{\mathrm{final}},
\end{equation}
where $L$ is the number of compact candidates whose initialization quality is
estimated, $S$ is the number of refinement trials per candidate,
$T_{\mathrm{move}}=O(nK)$ is the cost of one candidate-restricted improvement
step in the original space, and $T_{\mathrm{final}}\le B_\mu
T_{\mathrm{move}}$ is the final refinement. Because $L$, $S$, and $B$ are fixed
by the configuration, the cost of Just Initialize is a bounded multiple of a
single original-space sweep. The quantity it controls is therefore the number
of sweeps that downstream optimization must perform, which is exactly the
criterion introduced in Section~\ref{sec:refinement_initialization}.

\subsection{Validity of the Compact-Space Objective}
\label{app:theory_estimator}

Fix a compact candidate $\tilde\pi$ and let its recovered initialization be
drawn as $\pi_0\sim Q_M(\cdot\mid\tilde\pi)$, where $Q_M$ is the sampling rule of
Section~\ref{sec:methodology}. Write
\begin{equation}
Y(\tilde\pi,\xi)=C_I\bigl(\mathcal R_I(\pi_0;B,\xi)\bigr),
\qquad
J_B(\tilde\pi)=\mathbb E\bigl[Y(\tilde\pi,\xi)\bigr],
\end{equation}
where $\mathcal R_I(\pi_0;B,\xi)$ is the solution returned by the downstream
solver started at $\pi_0$ with budget $B$, and $\xi$ collects any randomness in
refinement. This is the compact-space counterpart of the initialization
quality functional of Section~\ref{sec:refinement_initialization}: a lower
$J_B$ means a more effective initialization.

\paragraph{Proposition 3 (Unbiased estimation and concentration).}
Let $S\ge1$ independent trials produce $Y_1,\dots,Y_S$ and let
$\widehat J_B(\tilde\pi)=\frac1S\sum_{s=1}^{S}Y_s$. If the trials are
identically distributed with finite mean $J_B(\tilde\pi)$ and variance
$\sigma^2(\tilde\pi)$, then
\begin{equation}
\mathbb E\bigl[\widehat J_B(\tilde\pi)\bigr]=J_B(\tilde\pi),
\qquad
\operatorname{Var}\bigl(\widehat J_B(\tilde\pi)\bigr)
=\frac{\sigma^2(\tilde\pi)}{S}.
\end{equation}
If in addition $Y_s\in[0,C_{\max}]$ almost surely, then for every
$\varepsilon>0$
\begin{equation}
\Pr\Bigl(\bigl|\widehat J_B(\tilde\pi)-J_B(\tilde\pi)\bigr|
\ge\varepsilon\Bigr)
\le 2\exp\!\left(-\frac{2S\varepsilon^2}{C_{\max}^2}\right).
\end{equation}

\paragraph{Proof.}
Linearity of expectation gives the first identity, and independence removes the
covariance terms, so
$\operatorname{Var}(\widehat J_B)=\frac1{S^2}\sum_s\operatorname{Var}(Y_s)
=\sigma^2/S$. The tail bound is Hoeffding's inequality applied to the
independent, bounded variables $Y_s$.
\hfill $\square$

\paragraph{Corollary 1 (High-probability candidate selection).}
Let $\mathcal C$ be the finite set of compact candidates evaluated by the
compact-space search and let $S\ge1$. For any $\alpha\in(0,1)$, set
\begin{equation}
\varepsilon
=
C_{\max}\sqrt{\frac{1}{2S}\ln\frac{2|\mathcal C|}{\alpha}}.
\end{equation}
With probability at least $1-\alpha$,
$\max_{\tilde\pi\in\mathcal C}|\widehat J_B(\tilde\pi)-J_B(\tilde\pi)|
\le\varepsilon$, and the selected candidate
$\widehat{\tilde\pi}\in\arg\min_{\tilde\pi\in\mathcal C}\widehat J_B(\tilde\pi)$
satisfies
\begin{equation}
J_B\bigl(\widehat{\tilde\pi}\bigr)
\le
\min_{\tilde\pi\in\mathcal C}J_B(\tilde\pi)+2\varepsilon .
\end{equation}

\paragraph{Proof.}
A union bound over $\mathcal C$ applied to
Eq.~(5) gives the uniform deviation event. On that event, with
$\tilde\pi^\star=\arg\min_{\tilde\pi\in\mathcal C}J_B(\tilde\pi)$,
\begin{equation}
J_B(\widehat{\tilde\pi})
\le\widehat J_B(\widehat{\tilde\pi})+\varepsilon
\le\widehat J_B(\tilde\pi^\star)+\varepsilon
\le J_B(\tilde\pi^\star)+2\varepsilon,
\end{equation}
where the second inequality uses the definition of $\widehat{\tilde\pi}$.
\hfill $\square$

Corollary~1 justifies the compact-space update of Eq.~(8): the candidate
returned by the search is, with high probability, within $2\varepsilon$ of the
best initialization quality in the evaluated set, and the deviation shrinks as
$1/\sqrt S$. When refinement is deterministic and recovery is concentrated on a
single candidate — the low-temperature setting used in the experiments —
$Y$ is deterministic, $\sigma^2=0$, a single trial evaluates $J_B$ exactly, and
the bound holds with $\varepsilon=0$.

\subsection{Decomposition of the Residual Error}
\label{app:theory_error}

Fix a compact structure $z\in\Omega(\tilde I)$ and let
\begin{equation}
\Omega_z=\{\pi\in\Omega(I):\Phi_s(\pi;M)=z\},
\qquad
C_z^*=\min_{\pi\in\Omega_z}C_I(\pi),
\end{equation}
with $\Omega_z\neq\emptyset$, and let $C_I^*=\min_{\pi\in\Omega(I)}C_I(\pi)$.
Both minima are attained because the feasible spaces are finite and the costs
are finite.

\paragraph{Proposition 4 (Error decomposition).}
For any recovered initialization $\pi_r\in\Omega_z$,
\begin{equation}
C_I(\pi_r)-C_I^*
=
\underbrace{\bigl(C_z^*-C_I^*\bigr)}_{\text{structural error}}
+
\underbrace{\bigl(C_I(\pi_r)-C_z^*\bigr)}_{\text{recovery error}},
\end{equation}
and both terms are nonnegative.

\paragraph{Proof.}
Adding and subtracting $C_z^*$ on the left gives the identity. Since
$\Omega_z\subseteq\Omega(I)$, we have $C_z^*\ge C_I^*$, and since
$\pi_r\in\Omega_z$, we have $C_I(\pi_r)\ge C_z^*$; both terms are therefore
nonnegative.
\hfill $\square$

\paragraph{Corollary 2 (Residual error after refinement).}
Let $\pi_f$ be the solution returned by the descent refinement started at
$\pi_r$. Since every accepted move strictly decreases the objective,
$C_I(\pi_f)\le C_I(\pi_r)$, and hence
\begin{equation}
C_I(\pi_f)-C_I^*
\le
\bigl(C_z^*-C_I^*\bigr)+\bigl(C_I(\pi_r)-C_z^*\bigr).
\end{equation}

\paragraph{Proof.}
The descent operator is non-increasing by definition of the acceptance rule, so
$C_I(\pi_f)\le C_I(\pi_r)$; substituting Eq.~(10) gives the bound.
\hfill $\square$

The two terms in Eq.~(10) separate the responsibilities of the two stages of
Just Initialize. The structural error depends only on which structure was
selected, and is therefore controlled by instance compression and by the
compact-space search; no amount of improvement within $\Omega_z$ can reduce it.
The recovery error depends only on how accurately the recovered
initialization instantiates the selected structure, and is the term that local
refinement absorbs. This is what licenses aggressive compression: the
compressed representation must carry the global structure that determines
$C_z^*$, while local detail is not required to be accurate at compression time,
because Corollary~2 charges it to the recovery term. It is also consistent with
the component ablation of Section~\ref{sec:ablation}, where replacing
compression degrades the final gap more than removing the compact-space search:
the former raises the structural term, the latter leaves it essentially
unchanged.

\paragraph{Remark 1 (Immediate cost does not determine initialization quality).}
$J_B$ is not a function of $C_I(\pi_0)$. Two initializations of equal objective
value can lie in different basins of the descent operator, so that from one the
operator reaches a strictly better local minimum within budget $B$ than from the
other. Comparing initializations by their immediate objective is therefore not
sound, which is why Section~\ref{sec:refinement_initialization} compares them by
the expected refined objective at a fixed solver and budget, and why the
comparisons of Section~\ref{sec:solver_initialization} and
Appendix~\ref{sec:appendix_greedy_starts} hold the refinement procedure fixed.
\section{Datasets and Parameters}
\label{app:datasets_parameters}

\subsection{Datasets}
The TSP collection contains 282 instances in three groups. The uniform group has 16 instances at each of 1K, 5K, 10K, 20K, 50K, and 100K nodes; the main comparison reports the 80 instances from 5K to 100K. The extended group follows the clustered, explosion, and implosion generators used by INViT~\citep{fang2024invit}; each distribution has 16 instances at 1K, 10K, and 100K nodes, for 144 instances in total. The third group contains 42 public benchmark instances, including TSPLIB95~\citep{reinelt1991tsplib} cases such as \texttt{pr2392}, \texttt{rl11849}, and \texttt{pla85900}, together with instances from the National TSP, VLSI, TSP Art, and USA collections. We retain their source coordinates and distance definitions: 39 use \texttt{EUC\_2D} and three use \texttt{CEIL\_2D}. The synthetic TSP groups use \texttt{EUC\_2D} after coordinate normalization and integer export.

The CVRP collection has 96 uniform instances, with 16 at each of 1K, 5K, 10K, 20K, 50K, and 100K customers; the main comparison reports the 80 instances from 5K to 100K. VRPTW and PCTSP each contain 72 uniform instances: 16 at each of 1K, 2K, 5K, and 10K customers and eight at 20K. These sizes count customers; each CVRP, VRPTW, and PCTSP instance also has one depot. The VRPTW generator first creates 16 candidates at 20K. The published set contains the eight lowest-reference-cost instances among the 11 candidates for which the reference search converged; the retained instances were renumbered from 01 to 08.

\subsection{Parameters}
Table~\ref{tab:dataset_generation} collects the generation settings. Here $n$ is the number of cities for TSP or customers for the other problems, and $i$ is the original generation index within a scale. Each seed initializes one instance independently. The selected VRPTW 20K instances retain their original seeds after renumbering. Synthetic coordinates lie in the unit square and are exported on a $10^6$ integer grid. For CVRP, VRPTW, and PCTSP, the canonical files retain the continuous coordinates and attributes; the corresponding solver files contain the integer export. The TSP exporter deterministically resolves coordinate collisions introduced by rounding.

\begin{table}[H]
\centering
\caption{Instance generation settings for the full collections. $U[a,b]$ denotes a continuous uniform distribution and $U_{\mathbb Z}[a,b]$ an integer uniform distribution.}
\label{tab:dataset_generation}
\footnotesize
\setlength{\tabcolsep}{3pt}
\begin{tabular}{@{}p{0.12\linewidth}p{0.22\linewidth}p{0.43\linewidth}p{0.17\linewidth}@{}}
\toprule
Family & Sizes $\times$ instances & Sampling and fixed settings & Generation seed \\
\midrule
TSP uniform & 1K, 5K, 10K, 20K, 50K, 100K $\times16$ & Independent uniform two-dimensional coordinates & $s_U(n,i)$ \\
TSP extended & 1K, 10K, 100K $\times16$ per distribution & Clustered, explosion, or implosion coordinates & $s_E(n,i)$ \\
CVRP & 1K, 5K, 10K, 20K, 50K, 100K $\times16$ & Depot/customers $\sim U[0,1]^2$; demand $\sim U_{\mathbb Z}[1,9]$ & $s_R(n,i)$ \\
VRPTW & 1K, 2K, 5K, 10K $\times16$; 20K $\times8$ & CVRP-style coordinates and demands; depot window $[0,3]$; service time $0.2$ & $s_R(n,i)$ \\
PCTSP & 1K, 2K, 5K, 10K $\times16$; 20K $\times8$ & Depot/customers $\sim U[0,1]^2$; prize $\sim U[0,4/n]$; penalty $\sim U[0,1.2/\sqrt n]$; quota $1$ & $s_R(n,i)$ \\
\bottomrule
\end{tabular}
\end{table}

The seed rules are $s_U(n,i)=420000+n+i$ for uniform TSP, $s_E(n,i)=b+n+i$ for extended TSP, and $s_R(n,i)=20260909+97n+1000003i$ for CVRP, VRPTW, and PCTSP. In $s_E$, $b$ is 520000 for clustered, 620000 for explosion, and 720000 for implosion.

For the extended TSP distributions, the clustered generator assigns nodes to three Gaussian clusters with center diversity 10. The explosion and implosion generators draw a center and a radius in $[0.1,0.5]$, then transform points inside the radius; explosion uses an exponential draw with rate 10, while implosion uses a normal draw. All generated TSP point sets are normalized before integer export. In the full CVRP collection, capacities are 250, 500, 1000, and 2000 for 1K, 5K, 10K, and 20K customers, respectively, and remain 2000 at 50K and 100K. VRPTW uses capacities 250, 500, 500, 1000, and 2000 at its five published scales. Its customer-window center is sampled between the travel time from the depot and the latest start allowing a direct return; the half-width is sampled uniformly from $[0.1,1.0]$ and clipped to the depot window. Instances are resampled until every customer admits a feasible depot--customer--depot trip.

The dataset parameters above describe instance generation; the following fixed values govern the frozen solvers used for the Ours results. Reference-value searches use their own budgets. The TSP solver grows path fragments toward a target length of 32, using 64 candidate neighbors per node and examining eight neighbors when extending a fragment. Its fragment graph retains 16 portal neighbors per fragment. The compact solver uses beam width 6 and at most eight coarse 2-opt passes. On the original tour, refinement uses 64 candidate neighbors and 16 iterated-search perturbations, selecting perturbation positions among the 80 longest edges. The initial local search allows up to 16 passes of 3-opt, and the subsequent Lin--Kernighan search has depth 3.

The CVRP solver uses a Clarke--Wright parameter of $\lambda=1.4$ and a fragment demand threshold of $0.8Q$, where $Q$ is vehicle capacity. It uses 32 candidate neighbors per customer and 16 portal neighbors per fragment. Compact-route improvement allows up to 200 inter-route relocate moves. Route-level 2-opt examines 16 candidate neighbors and accepts at most ten moves per route during compression and restoration, followed by at most one move per route in the final refinement stage. The final 2-opt* search examines eight candidate neighbors and accepts at most four moves.

\section{Extended Experiments}
\raggedbottom
\label{app:Extended-Experiments}
\subsection{Non-uniform TSP instances}

\paragraph{Experiment Design.}
To evaluate the generalization ability of Just Initialize (JI), we further conduct experiments on non-uniform TSP instances with three challenging spatial distributions: clustered, explosion, and implosion.
Each distribution contains 16 instances at 1K, 10K, and 100K nodes.
Unlike learning-based approaches that rely on training data, JI is completely training-free and does not require distribution alignment between training and testing instances.
This experiment evaluates whether JI can provide effective initializations under diverse instance distributions.

\begin{table}[ht]
\centering
\caption{TSP results on clustered distributions (16 instances per size). 
Bold marks the lowest gap and solve time at each size, excluding Concorde; gray highlights Ours.}
\label{tab:tsp_clustered}
\small
\setlength{\tabcolsep}{3pt}
\renewcommand{\arraystretch}{0.95}
\begin{tabular}{@{}l|ccc|ccc|ccc@{}}
\toprule
Method & \multicolumn{3}{c|}{1K} & \multicolumn{3}{c|}{10K} & \multicolumn{3}{c}{100K} \\
& Obj.$\downarrow$ & Gap (\%) & Time
& Obj.$\downarrow$ & Gap (\%) & Time
& Obj.$\downarrow$ & Gap (\%) & Time \\
\midrule

LKH-3 (2.0\,h) 
& 14.22 & -- & --
& 40.67 & -- & --
& 120.22 & -- & -- \\

Concorde 
& 14.22 & 0.00 & 1.7\,min
& 40.86 & 0.48 & 3.0\,min
& 121.35 & 0.94 & 3.0\,min \\

\midrule

POMO (no aug.) 
& 22.32 & 56.96 & 0.6\,s
& 109.77 & 169.91 & 6.5\,min
& \multicolumn{3}{c}{OOM} \\

POMO ($\times 8$ aug.)
& 20.96 & 47.41 & 3.1\,s
& \multicolumn{3}{c|}{OOM}
& \multicolumn{3}{c}{OOM} \\

LEHD (greedy)
& 15.51 & 9.07 & 4.5\,s
& 54.95 & 35.11 & 8.0\,min
& \multicolumn{3}{c}{OOM} \\

INViT
& 15.23 & 7.09 & 8.9\,s
& 44.10 & 8.43 & 1.5\,min
& 129.49 & 7.71 & 18.6\,min \\

SIL
& 15.35 & 7.97 & 3.9\,s
& 86.58 & 112.89 & 12.0\,min
& \multicolumn{3}{c}{OOM} \\

L2C-Insert ($I=1000$)
& 14.59 & \textbf{2.59} & 57.7\,s
& 41.65 & 4.15 & 1.0\,min
& 126.81 & 5.49 & 1.6\,min \\

\midrule

DIFUSCO
& 16.22 & 14.04 & 11.8\,s
& 45.85 & 12.76 & 34.3\,s
& \multicolumn{3}{c}{OOM} \\

GLOP
& 15.11 & 6.31 & 1.4\,s
& 43.47 & 6.89 & 1.8\,s
& 128.67 & 7.03 & 1.2\,min \\

UDC
& 14.62 & 2.87 & 3.4\,s
& 42.37 & 4.20 & 3.7\,s
& 125.61 & 4.48 & 1.2\,min \\

H-TSP
& 15.41 & 8.41 & 1.2\,s
& 44.46 & 9.31 & 9.3\,s
& 130.92 & 8.90 & 1.7\,min \\

\midrule

TTPL
& 14.80 & 4.09 & 6.9\,s
& 42.04 & 3.74 & 1.1\,min
& 124.22 & 3.33 & 11.5\,min \\

\midrule

DeepACO ($T=10$)
& 14.73 & 3.58 & 3.8\,min
& \multicolumn{3}{c|}{OOM}
& \multicolumn{3}{c}{OOM} \\

DeepACO ($T=2$)
& 14.87 & 4.61 & 53.2\,s
& \multicolumn{3}{c|}{OOM}
& \multicolumn{3}{c}{OOM} \\

\midrule

\rowcolor{gray!35}
Ours
& 14.63 & 2.89 & \textbf{0.1\,s}
& 42.14 & \textbf{3.62} & \textbf{2.0\,s}
& 124.57 & \textbf{3.11} & \textbf{37.2\,s} \\

\bottomrule
\end{tabular}
\end{table}

\begin{table}[ht]
\centering
\caption{TSP results on explosion distributions (16 instances per size). 
Bold marks the lowest gap and solve time at each size, excluding Concorde; gray highlights Ours.}
\label{tab:tsp_explosion}
\small
\setlength{\tabcolsep}{3pt}
\renewcommand{\arraystretch}{0.95}
\begin{tabular}{@{}l|ccc|ccc|ccc@{}}
\toprule
Method & \multicolumn{3}{c|}{1K} & \multicolumn{3}{c|}{10K} & \multicolumn{3}{c}{100K} \\
& Obj.$\downarrow$ & Gap (\%) & Time
& Obj.$\downarrow$ & Gap (\%) & Time
& Obj.$\downarrow$ & Gap (\%) & Time \\
\midrule

LKH-3 (2.0\,h)
& 17.01 & -- & --
& 40.26 & -- & --
& 108.17 & -- & -- \\

Concorde
& 17.01 & 0.00 & 2.4\,min
& 40.43 & 0.42 & 3.0\,min
& 109.33 & 1.07 & 3.0\,min \\

\midrule

POMO (no aug.)
& 25.91 & 52.34 & 0.6\,s
& 117.39 & 191.57 & 6.5\,min
& \multicolumn{3}{c}{OOM} \\

POMO ($\times 8$ aug.)
& 24.75 & 45.48 & 3.1\,s
& \multicolumn{3}{c|}{OOM}
& \multicolumn{3}{c}{OOM} \\

LEHD (greedy)
& 17.91 & 5.32 & 4.1\,s
& 53.09 & 31.88 & 8.0\,min
& \multicolumn{3}{c}{OOM} \\

INViT
& 18.30 & 7.57 & 8.9\,s
& 44.36 & 10.18 & 1.5\,min
& 117.23 & 8.38 & 18.5\,min \\

SIL
& 17.57 & 3.30 & 3.9\,s
& 99.48 & 147.09 & 12.0\,min
& \multicolumn{3}{c}{OOM} \\

L2C-Insert ($I=1000$)
& 17.24 & \textbf{1.40} & 57.2\,s
& 41.32 & 4.23 & 1.0\,min
& 114.60 & 5.94 & 1.6\,min \\

\midrule

DIFUSCO
& 18.98 & 11.60 & 3.2\,s
& 45.69 & 13.49 & 34.3\,s
& \multicolumn{3}{c}{OOM} \\

GLOP
& 17.95 & 5.53 & 1.4\,s
& 43.01 & 6.84 & 1.8\,s
& 115.75 & 7.01 & 1.2\,min \\

UDC
& 17.47 & 2.64 & 3.4\,s
& 41.96 & 4.22 & 3.8\,s
& 113.11 & 4.57 & 1.2\,min \\

H-TSP
& 18.56 & 9.11 & 1.3\,s
& 44.19 & 9.75 & 9.7\,s
& 119.52 & 10.49 & 1.7\,min \\

\midrule

TTPL
& 18.01 & 5.87 & 6.9\,s
& 42.73 & 6.13 & 1.1\,min
& 113.26 & 4.71 & 11.6\,min \\

\midrule

DeepACO ($T=10$)
& 17.57 & 3.28 & 3.7\,min
& \multicolumn{3}{c|}{OOM}
& \multicolumn{3}{c}{OOM} \\

DeepACO ($T=2$)
& 17.68 & 3.96 & 50.9\,s
& \multicolumn{3}{c|}{OOM}
& \multicolumn{3}{c}{OOM} \\

\midrule

\rowcolor{gray!35}
Ours
& 17.49 & 2.83 & \textbf{0.1\,s}
& 41.77 & \textbf{3.75} & \textbf{1.9\,s}
& \textbf{112.17} & \textbf{3.76} & \textbf{36.7\,s} \\

\bottomrule
\end{tabular}
\end{table}
\begin{table}[ht]
\centering
\caption{TSP results on implosion distributions (16 instances per size). 
Bold marks the lowest gap and solve time at each size, excluding Concorde; gray highlights Ours.}
\label{tab:tsp_implosion}
\small
\setlength{\tabcolsep}{3pt}
\renewcommand{\arraystretch}{0.95}
\begin{tabular}{@{}l|ccc|ccc|ccc@{}}
\toprule
Method & \multicolumn{3}{c|}{1K} & \multicolumn{3}{c|}{10K} & \multicolumn{3}{c}{100K} \\
& Obj.$\downarrow$ & Gap (\%) & Time
& Obj.$\downarrow$ & Gap (\%) & Time
& Obj.$\downarrow$ & Gap (\%) & Time \\
\midrule

LKH-3 (2.0\,h)
& 20.29 & -- & --
& 63.25 & -- & --
& 189.57 & -- & -- \\

Concorde
& 20.29 & 0.00 & 1.7\,min
& 63.53 & 0.44 & 3.0\,min
& 191.64 & 1.10 & 3.0\,min \\

\midrule

POMO (no aug.)
& 28.85 & 42.20 & 0.6\,s
& 119.39 & 88.76 & 6.5\,min
& \multicolumn{3}{c}{OOM} \\

POMO ($\times 8$ aug.)
& 28.39 & 39.94 & 3.1\,s
& \multicolumn{3}{c|}{OOM}
& \multicolumn{3}{c}{OOM} \\

LEHD (greedy)
& 21.05 & 3.75 & 4.2\,s
& 81.98 & 29.61 & 8.0\,min
& \multicolumn{3}{c}{OOM} \\

INViT
& 21.50 & 5.94 & 8.9\,s
& 68.00 & 7.53 & 1.5\,min
& 203.77 & 7.49 & 18.5\,min \\

SIL
& 21.26 & 4.78 & 4.0\,s
& 90.25 & 42.68 & 12.0\,min
& \multicolumn{3}{c}{OOM} \\

L2C-Insert ($I=1000$)
& 20.45 & \textbf{0.78} & 57.3\,s
& 64.75 & \textbf{2.39} & 1.0\,min
& 199.28 & 5.12 & 1.6\,min \\

\midrule

DIFUSCO
& 22.49 & 10.85 & 3.2\,s
& 70.68 & 11.75 & 34.3\,s
& \multicolumn{3}{c}{OOM} \\

GLOP
& 21.33 & 5.09 & 1.4\,s
& 67.37 & 6.50 & 1.9\,s
& 202.59 & 6.87 & 1.2\,min \\

UDC
& 20.86 & 2.76 & 3.4\,s
& 65.75 & 3.96 & 3.8\,s
& 197.83 & 4.35 & 1.2\,min \\

H-TSP
& 21.92 & 7.99 & 1.2\,s
& 68.98 & 9.06 & 9.3\,s
& 208.96 & 10.23 & 1.7\,min \\

\midrule

TTPL
& 21.08 & 3.88 & 6.9\,s
& 65.99 & 4.33 & 1.1\,min
& 196.36 & 3.84 & 11.6\,min \\

\midrule

DeepACO ($T=10$)
& 20.91 & 3.02 & 3.9\,min
& \multicolumn{3}{c|}{OOM}
& \multicolumn{3}{c}{OOM} \\

DeepACO ($T=2$)
& 21.01 & 3.55 & 53.4\,s
& \multicolumn{3}{c|}{OOM}
& \multicolumn{3}{c}{OOM} \\

\midrule

\rowcolor{gray!35}
Ours
& 20.88 & 2.88 & \textbf{0.1\,s}
& 65.48 & 3.52 & \textbf{1.9\,s}
& 196.66 & \textbf{3.74} & \textbf{38.5\,s} \\

\bottomrule
\end{tabular}
\end{table}

\paragraph{Results.}
As shown in Tables~\ref{tab:tsp_clustered}, 
\ref{tab:tsp_explosion}, and 
\ref{tab:tsp_implosion},
JI consistently achieves competitive solution quality across all three distributions and different scales, while maintaining substantially lower computational costs.
For 10K-node instances, JI obtains low gaps within only a few seconds, and for 100K-node instances, it remains efficient with runtimes within tens of seconds.
These results demonstrate that JI can be effectively combined with downstream refinement solvers without additional training or distribution-specific adaptation.

\paragraph{Analysis.}
The stable performance across clustered, explosion, and implosion distributions demonstrates the strong generalization capability of JI.
Although learning-based approaches can exploit distribution-specific patterns through training, their effectiveness may depend on the similarity between training and testing distributions.
In contrast, JI does not require such distribution alignment; it directly identifies promising regions from each instance through geometric compression and generates solver-compatible initializations.
Therefore, JI provides a distribution-agnostic initialization mechanism, enabling efficient refinement across diverse routing scenarios.
\subsection{TSPLIB95 instances}

\paragraph{Experiment Design.}
To further evaluate the generalization ability of Just Initialize (JI) on real-world routing instances, we conduct experiments on 32 TSPLIB95 instances with different scales, ranging from 1K to 85K nodes.
These instances contain diverse spatial structures and are widely used as standard benchmarks for evaluating large-scale TSP solvers.
JI is directly applied without additional training or instance-specific adaptation, demonstrating its ability to generate effective initializations for unseen routing scenarios.

\paragraph{Results.}
Table~\ref{tab:tsp_tsplib95} reports the relative optimality gaps on TSPLIB95 instances.
JI achieves the lowest average gap among all compared methods, reducing the average gap to 3.54\%.
Moreover, JI obtains the best-reported gap on 19 out of 32 instances, while successfully solving all instances in the benchmark.
Compared with existing methods, JI maintains consistently low gaps across different scales, including large instances with tens of thousands of nodes.

\paragraph{Analysis.}
The results demonstrate that JI can generalize beyond synthetic distributions to diverse real-world routing instances.
Unlike learning-based methods that rely on training distributions to capture instance patterns, JI does not require such distribution alignment.
Instead, JI extracts the underlying geometric structure of each instance through compression and provides a promising initialization for downstream optimization.
This enables JI to remain effective across different instance scales and spatial characteristics, validating the robustness of the proposed initialization mechanism.

\begingroup
\begin{table}[H]
\centering
\caption{Relative gaps (\%) on TSPLIB95 instances. Bold marks the lowest reported gap in each row; gray highlights Ours. OOM marks out-of-memory on a 24\,GB GPU.}
\label{tab:tsp_tsplib95}
\small
\setlength{\tabcolsep}{4pt}
\renewcommand{\arraystretch}{0.95}
\begin{tabular}{l r|rrrrr>{\columncolor{gray!35}}c}
\toprule
Instance & Scale & LEHD & TTPL & H-TSP & UDC & GLOP & JI+Refine (Ours) \\
\midrule
pr1002 & 1,002 & 4.43\% & \textbf{0.73\%} & 7.19\% & 2.62\% & 5.21\% & 4.41\% \\
u1060 & 1,060 & 10.01\% & 2.54\% & 10.84\% & 3.55\% & 4.82\% & \textbf{2.08}\% \\
vm1084 & 1,084 & 5.42\% & \textbf{1.44\%} & 12.70\% & 3.95\% & 5.61\% & 3.29\% \\
pcb1173 & 1,173 & 7.95\% & \textbf{1.72\%} & 8.51\% & 5.23\% & 7.48\% & 4.36\% \\
d1291 & 1,291 & 13.46\% & 6.22\% & 16.11\% & 8.32\% & 9.59\% & \textbf{2.47\%} \\
rl1304 & 1,304 & 8.14\% & 4.16\% & 14.67\% & 6.94\% & 10.00\% & \textbf{2.82\%} \\
rl1323 & 1,323 & 9.27\% & 3.89\% & 16.05\% & 5.28\% & 11.58\% & \textbf{2.53\%} \\
nrw1379 & 1,379 & 15.43\% & \textbf{0.98\%} & 6.47\% & 2.20\% & 5.08\% & 3.73\% \\
fl1400 & 1,400 & 18.10\% & 16.99\% & 19.64\% & 4.65\% & 5.50\% & \textbf{3.48\%} \\
u1432 & 1,432 & 7.93\% & \textbf{2.15\%} & 8.95\% & 2.65\% & 5.83\% & 4.88\% \\
fl1577 & 1,577 & 14.98\% & 11.01\% & 23.07\% & 7.52\% & 9.69\% & \textbf{6.66\%} \\
d1655 & 1,655 & 13.68\% & 6.46\% & 11.69\% & 4.23\% & 6.93\% & \textbf{2.88\%} \\
vm1748 & 1,748 & 10.11\% & 8.25\% & 10.78\% & 3.95\% & 5.51\% & \textbf{2.77\%} \\
u1817 & 1,817 & 8.98\% & 4.18\% & 11.94\% & 5.43\% & 10.52\% & \textbf{3.24\%} \\
rl1889 & 1,889 & 7.50\% & 4.91\% & 12.40\% & 7.57\% & 11.31\% & \textbf{2.86\%} \\
d2103 & 2,103 & 11.97\% & 8.01\% & 15.80\% & 8.80\% & 14.91\% & \textbf{2.50\%} \\
u2152 & 2,152 & 9.27\% & \textbf{2.47\%} & 14.81\% & 5.85\% & 9.68\% & 4.10\% \\
u2319 & 2,319 & 4.14\% & \textbf{0.22\%} & 2.20\% & 1.04\% & 2.19\% & 2.26\% \\
pr2392 & 2,392 & 12.31\% & \textbf{2.85\%} & 18.37\% & 4.86\% & 7.14\% & 2.88\% \\
pcb3038 & 3,038 & 13.37\% & 5.26\% & 8.45\% & 4.71\% & 7.98\% & \textbf{3.88\%} \\
fl3795 & 3,795 & 17.51\% & 34.75\% & 19.09\% & 5.96\% & 9.51\% & \textbf{5.65\%} \\
fnl4461 & 4,461 & 18.95\% & \textbf{2.39\%} & 6.44\% & 3.14\% & 5.97\% & 3.17\% \\
rl5915 & 5,915 & 24.18\% & 4.57\% & 16.11\% & 9.90\% & 13.44\% & \textbf{2.78\%} \\
rl5934 & 5,934 & 24.11\% & 8.31\% & 16.39\% & 8.33\% & 11.64\% & \textbf{3.54\%} \\
pla7397 & 7,397 & 40.94\% & 8.61\% & 10.74\% & 6.67\% & 7.73\% & \textbf{3.25\%} \\
rl11849 & 11,849 & 37.51\% & 5.27\% & 15.14\% & 7.85\% & 11.59\% & \textbf{3.89\%} \\
usa13509 & 13,509 & 70.17\% & 3.85\% & 10.11\% & 4.44\% & 7.16\% & \textbf{3.83}\% \\
brd14051 & 14,051 & OOM & 4.48\% & 9.46\% & \textbf{3.52\%} & 5.78\% & 3.67\% \\
d15112 & 15,112 & OOM & \textbf{3.20\%} & 7.99\% & 3.51\% & 6.06\% & 3.41\% \\
d18512 & 18,512 & OOM & \textbf{2.28\%} & 6.88\% & 3.65\% & 6.48\% & 3.34\% \\
pla33810 & 33,810 & OOM & \textbf{4.10\%} & 10.23\% & 8.31\% & 10.23\% & 4.77\% \\
pla85900 & 85,900 & OOM & 4.95\% & 9.91\% & 6.65\% & 9.19\% & \textbf{3.88\%} \\
\midrule
Solved\# &  & 27/32 & 32/32 & 32/32 & 32/32 & 32/32 & 32/32 \\
Best\# & & 0/32 & 12/32 & 0/32 & 1/32 & 0/32 & \textbf{19/32} \\
Avg.\ gap &  & 16.29\% & 5.66\% & 12.16\% & 5.35\% & 8.17\% & \textbf{3.54}\% \\
\bottomrule
\end{tabular}
\end{table}
\endgroup

\subsection{CVRPLIB instances}
\paragraph{Experiment Design.}
To further evaluate the applicability of Just Initialize (JI) beyond TSP, we conduct experiments on CVRPLIB instances with diverse scales and customer distributions.
The benchmark contains 11 instances ranging from 1K to 30K customers, where additional capacity constraints are introduced compared with TSP.
JI is directly applied without additional training or problem-specific adaptation, evaluating whether the proposed initialization mechanism can generalize to different routing formulations.

\paragraph{Results.}
Table~\ref{tab:cvrplib} reports the relative gaps on CVRPLIB instances.
JI successfully solves all 11 instances and achieves the lowest average gap among all compared methods, reaching an average gap of 7.77\%.
Moreover, JI obtains the best-reported gap on 8 out of 11 instances, demonstrating consistent performance across different problem scales.
These results show that JI maintains competitive solution quality while remaining applicable to large-scale vehicle routing problems with up to 30K customers.

\paragraph{Analysis.}
The results demonstrate that JI can generalize from TSP to more complex vehicle routing scenarios with additional constraints.
Unlike learning-based approaches that require problem-specific training to capture routing patterns, JI does not require such distribution alignment.
Instead, JI extracts the geometric structure of each instance through compression and generates effective initializations for downstream optimization.
This indicates that the proposed initialization mechanism is not limited to a specific routing formulation, but provides a general strategy for improving large-scale combinatorial optimization.

\begingroup
\begin{table}[H]
\centering
\caption{Relative gaps (\%) on CVRPLIB instances. Bold marks the lowest reported gap in each row; gray highlights Ours.}
\label{tab:cvrplib}
\small
\setlength{\tabcolsep}{4pt}
\renewcommand{\arraystretch}{0.95}
\begin{tabular}{l r|rrrrrrr>{\columncolor{gray!35}}r}
\toprule
Instance & Scale & HGS & OR-Tools & LEHD & L2C & TTPL & UDC & GLOP & Ours \\
\midrule
X-n1001-k43 & 1,000 & \textbf{3.01\%} & 9.59\% & 7.55\% & 7.53\% & 6.01\% & 6.49\% & 88.26\% & 6.10\% \\
Leuven1 & 3,000 & 5.42\% & 5.42\% & 16.36\% & 7.07\% & 6.48\% & 9.64\% & 105.31\% & \textbf{5.35\%} \\
Leuven2 & 4,000 & \textbf{9.00\%} & 12.36\% & 33.12\% & 25.76\% & 24.06\% & 44.00\% & 45.29\% & 12.52\% \\
Antwerp1 & 6,000 & 5.49\% & 7.25\% & 15.17\% & 6.23\% & 6.69\% & 9.59\% & 130.39\% & \textbf{4.47\%} \\
Antwerp2 & 7,000 & 11.27\% & 12.41\% & 21.36\% & 20.00\% & 13.54\% & 21.08\% & 63.61\% & \textbf{9.56\%} \\
Ghent1 & 10,000 & 6.74\% & 7.11\% & 27.64\% & 11.15\% & 10.64\% & 12.36\% & 146.15\% & \textbf{3.96\%} \\
Ghent2 & 11,000 & \textbf{10.43\%} & 14.92\% & 39.85\% & 30.34\% & 16.08\% & 184.76\% & 45.79\% & 11.36\% \\
Brussels1 & 15,000 & 9.54\% & 9.89\% & OOM & 15.41\% & 9.14\% & 17.13\% & 139.73\% & \textbf{6.17\%} \\
Brussels2 & 16,000 & 12.52\% & 16.28\% & OOM & 29.87\% & 15.03\% & 71.11\% & 55.88\% & \textbf{10.80\%} \\
Flanders1 & 20,000 & 6.43\% & 6.47\% & OOM & 18.07\% & 10.05\% & 13.91\% & 92.53\% & \textbf{4.31\%} \\
Flanders2 & 30,000 & 14.55\% & OOT & OOM & 45.81\% & 28.43\% & 309.02\% & 45.60\% & \textbf{10.87\%} \\
\midrule
Solved\# & & 11/11 & 10/11 & 7/11 & 11/11 & 11/11 & 11/11 & 11/11 & 11/11 \\
Best\# & & 3/11 & 0/11 & 0/11 & 0/11 & 0/11 & 0/11 & 0/11 & \textbf{8/11} \\
Avg.\ gap & & 8.58\% & 10.17\% & 23.01\% & 19.75\% & 13.29\% & 63.55\% & 87.14\% & \textbf{7.77}\% \\
\bottomrule
\end{tabular}
\end{table}
\endgroup

\subsection{Case Study}

\paragraph{Experiment Design.}
To investigate how the compact-space solve stage affects the quality of the generated initialization, 
we perform a case study on a 100K-node uniform TSP instance.
During the compact-space optimization process, we sample 24 intermediate solve states with different optimization budgets.
Each state is independently recovered to the original space and refined using the same refinement procedure and budget.
The resulting solutions are evaluated by their final GAP, allowing us to analyze whether later compact-space states provide better initialization quality.

\begin{table}[ht]
\centering
\caption{Intermediate solve states on the 100K uniform TSP instance.
$x$ denotes cumulative compact-space 2-opt evaluations (millions), and GAP is measured after recovery and refinement.}
\label{tab:solve_states}
\small
\setlength{\tabcolsep}{4pt}
\renewcommand{\arraystretch}{1.15}

\resizebox{\linewidth}{!}{
\begin{tabular}{@{}l*{12}{r}@{}}
\toprule
Point & 1 & 2 & 3 & 4 & 5 & 6 & 7 & 8 & 9 & 10 & 11 & 12 \\
\midrule
$x$ 
& 0.00 & 5.13 & 23.47 & 24.95 & 27.88 & 44.78 
& 48.84 & 51.61 & 51.76 & 71.21 & 74.72 & 76.97 \\

GAP (\%)
& 3.684 & 3.684 & 3.674 & 3.671 & 3.683 & 3.662
& 3.653 & 3.645 & 3.657 & 3.661 & 3.643 & 3.646 \\

\midrule

Point & 13 & 14 & 15 & 16 & 17 & 18 & 19 & 20 & 21 & 22 & 23 & 24 \\

$x$
& 91.20 & 93.77 & 101.18 & 102.97 & 103.52 & 104.91
& 107.66 & 111.03 & 123.94 & 146.85 & 151.44 & 162.71 \\

GAP (\%)
& 3.631 & 3.641 & 3.648 & 3.634 & 3.647 & 3.657
& 3.652 & 3.638 & 3.616 & 3.597 & 3.585 & 3.564 \\

\bottomrule
\end{tabular}
}
\end{table}

\begin{figure}[H]
\centering
\includegraphics[width=1.0\linewidth]{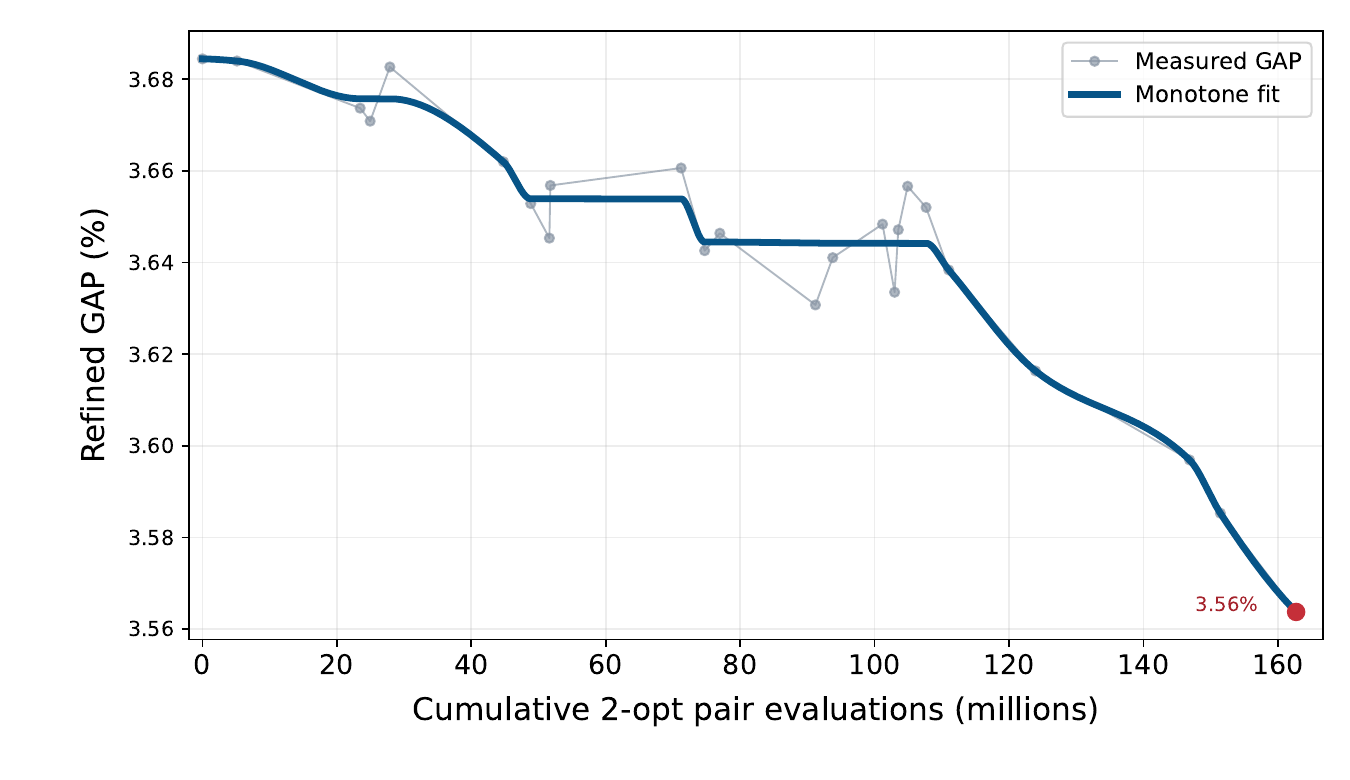}
\caption{Solve-state convergence of JI on the 100K uniform TSP instance. 
Gray markers denote measured GAP values, and the curve shows the overall decreasing trend.}
\label{fig:solve_convergence}
\end{figure}

\paragraph{Results.}
As shown in Table~\ref{tab:solve_states} and Figure~\ref{fig:solve_convergence}, 
the refined GAP consistently decreases as the compact-space optimization progresses.
The GAP decreases from 3.684\% at the initial checkpoint to 3.564\% at the final checkpoint, despite small local fluctuations during intermediate stages.
This demonstrates that additional optimization in the compact space leads to increasingly effective initializations for downstream refinement.

\paragraph{Analysis.}
The observed trajectory validates the motivation of JI: the compact-space solve stage is not directly optimizing the final solution, but identifying regions with stronger refinement potential in the original solution space.
Although compact-space improvements do not guarantee monotonic decreases after recovery, later solve states generally produce better refined solutions under the same refinement budget.
These results indicate that JI effectively guides the search toward more favorable regions before applying expensive fine-grained optimization.
\subsection{Convergence from Matched Starting Gaps}
\label{sec:appendix_matched_gap}
\paragraph{Experiment Design.}
To evaluate whether JI provides better refinement potential beyond simply achieving a lower initial GAP, 
we conduct a matched initialization study.
Three representative initialization strategies, including randomized nearest-neighbor (RNN), Hilbert space-filling-curve ordering (Hilbert), and minimum-spanning-tree traversal (MST), are selected with initial GAPs matched to JI within 0.05 percentage points.
All initializations undergo the same 64-neighbor 2-opt refinement procedure, and their final GAPs and refinement times are compared.

\begin{table}[H]
\centering
\caption{Final GAP and refinement time under matched initializations.}
\label{tab:matched_gap_refinement}
\vspace{4pt}
\small
\setlength{\tabcolsep}{4pt}
\renewcommand{\arraystretch}{0.95}
\begin{tabular}{@{}lrrrrrrrr@{}}
\toprule
& \multicolumn{4}{c}{Final relative gap (\%)} & \multicolumn{4}{c}{Search time (s)} \\
\cmidrule(lr){2-5}\cmidrule(l){6-9}
Scale & JI & RNN & Hilbert & MST & JI & RNN & Hilbert & MST \\
\midrule
1K   & \textbf{8.6338} & 10.7757 & 12.3769 & 10.5459 & 0.016 & 0.015 & 0.024 & 0.022 \\
5K   & \textbf{9.2353} & 11.3989 & 13.4103 & 11.0869 & 0.156 & 0.150 & 0.194 & 0.198 \\
10K  & \textbf{9.2395} & 11.2762 & 13.4725 & 11.1091 & 0.092 & 0.092 & 0.180 & 0.169 \\
20K  & \textbf{9.2093} & 11.3263 & 13.6263 & 11.1296 & 0.224 & 0.211 & 0.441 & 0.409 \\
50K  & \textbf{9.2136} & 11.1755 & 13.5666 & 11.1678 & 0.792 & 0.836 & 1.843 & 1.394 \\
100K & \textbf{9.1038} & 11.0169 & 13.5540 & 11.1422 & 3.276 & 3.104 & 6.338 & 4.274 \\
\bottomrule
\end{tabular}
\vspace{-8pt}
\end{table}

\begin{figure}[H]
\centering
\includegraphics[width=\linewidth]{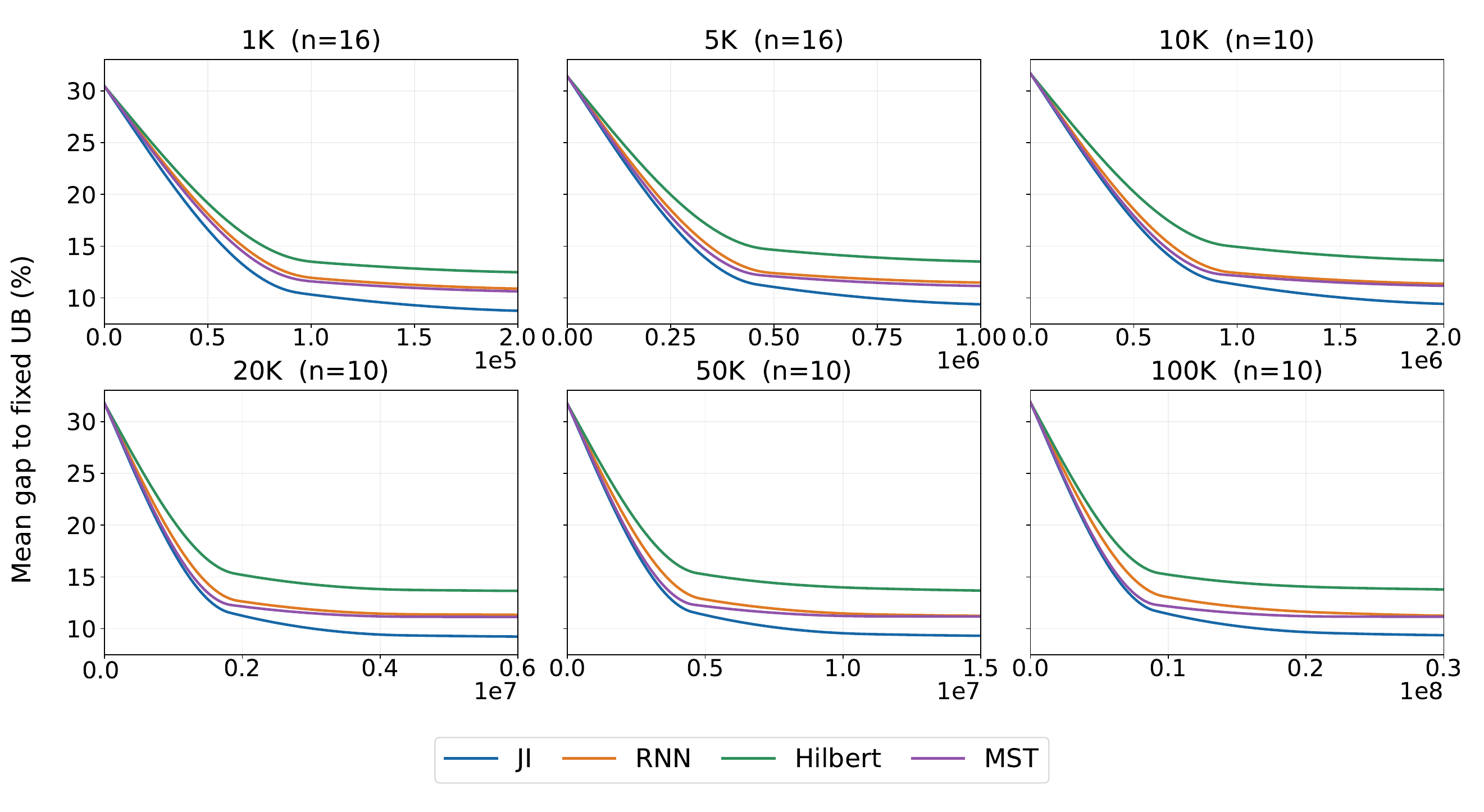}
\caption{GAP convergence trajectories during 2-opt refinement under matched initializations across six TSP scales.}
\label{fig:matched_gap_convergence}
\end{figure}
\paragraph{Results.}
Table~\ref{tab:matched_gap_refinement} and Figure~\ref{fig:matched_gap_convergence}report the refinement performance under matched initial GAPs.
JI consistently achieves lower final GAPs across all six scales while using comparable refinement procedures.
The convergence trajectories further show that JI maintains a faster improvement trend during optimization and reaches better final solutions than alternative initializations.

\paragraph{Analysis.}
The matched-gap comparison demonstrates that the effectiveness of JI does not come from simply providing a better initial objective value.
Instead, JI generates initial solutions located in regions with stronger downstream refinement potential.
Under comparable starting quality and identical optimization procedures, JI enables local search to reach better solutions, supporting our hypothesis that initialization quality should be evaluated by refinement effectiveness rather than immediate solution quality.

\subsection{Greedy Starts and Search Efficiency}
\label{sec:appendix_greedy_starts}

\paragraph{Experiment Design.}
To evaluate whether JI improves optimization efficiency through better initialization, 
we compare JI with a nearest-neighbor greedy initialization on 96 uniform TSP instances across six scales.
Both initializations are refined using the same deterministic 64-neighbor 2-opt procedure.
We evaluate the final solution quality, refinement time, convergence trajectory, and the number of candidate evaluations required to reach a target GAP of 10\%.
This controlled comparison isolates the effect of initialization quality from the optimization procedure.
\begin{table}[ht]
\centering
\begin{minipage}{0.48\linewidth}
\centering
\caption{Final GAP and 2-opt refinement time using greedy and JI initializations across different TSP scales.}
\label{tab:greedy_final}
\small
\setlength{\tabcolsep}{4pt}
\begin{tabular}{c|cc|cc}
\toprule
\multirow{2}{*}{Scale}
& \multicolumn{2}{c|}{Greedy}
& \multicolumn{2}{c}{JI} \\
\cmidrule(lr){2-3}
\cmidrule(l){4-5}
& GAP(\%) & Time(s)
& GAP(\%) & Time(s) \\
\midrule
1K
& 8.4431 & 0.012
& 7.5405 & 0.012 \\
5K
& 8.0722 & 0.072
& 7.7279 & 0.069 \\
10K
& 8.2994 & 0.161
& 7.6362 & 0.145 \\
20K
& 7.9563 & 0.373
& 7.6374 & 0.330 \\
50K
& 7.9596 & 1.068
& 7.6603 & 0.785 \\
100K
& 7.7531 & 2.826
& 7.6640 & 1.879 \\
\bottomrule
\end{tabular}
\end{minipage}
\hfill
\begin{minipage}{0.48\linewidth}
\centering
\caption{Search budgets for reaching a 10\% relative GAP under different initializations.}
\label{tab:search_budget}
\small
\setlength{\tabcolsep}{4pt}
\begin{tabular}{c|ccc}
\toprule
Scale
& JI
& Greedy
& Saving(\%) \\
\midrule
1K
& 60,532
& 100,415
& 39.72 \\
5K
& 308,912
& 434,586
& 28.92 \\
10K
& 585,714
& 1,000,608
& 41.46 \\
20K
& 1,201,814
& 1,837,954
& 34.61 \\
50K
& 2,972,090
& 4,640,830
& 35.96 \\
100K
& 5,940,941
& 8,894,603
& 33.21 \\
\bottomrule
\end{tabular}
\end{minipage}
\end{table}

\begin{figure}[H]
\centering
\includegraphics[width=\linewidth]{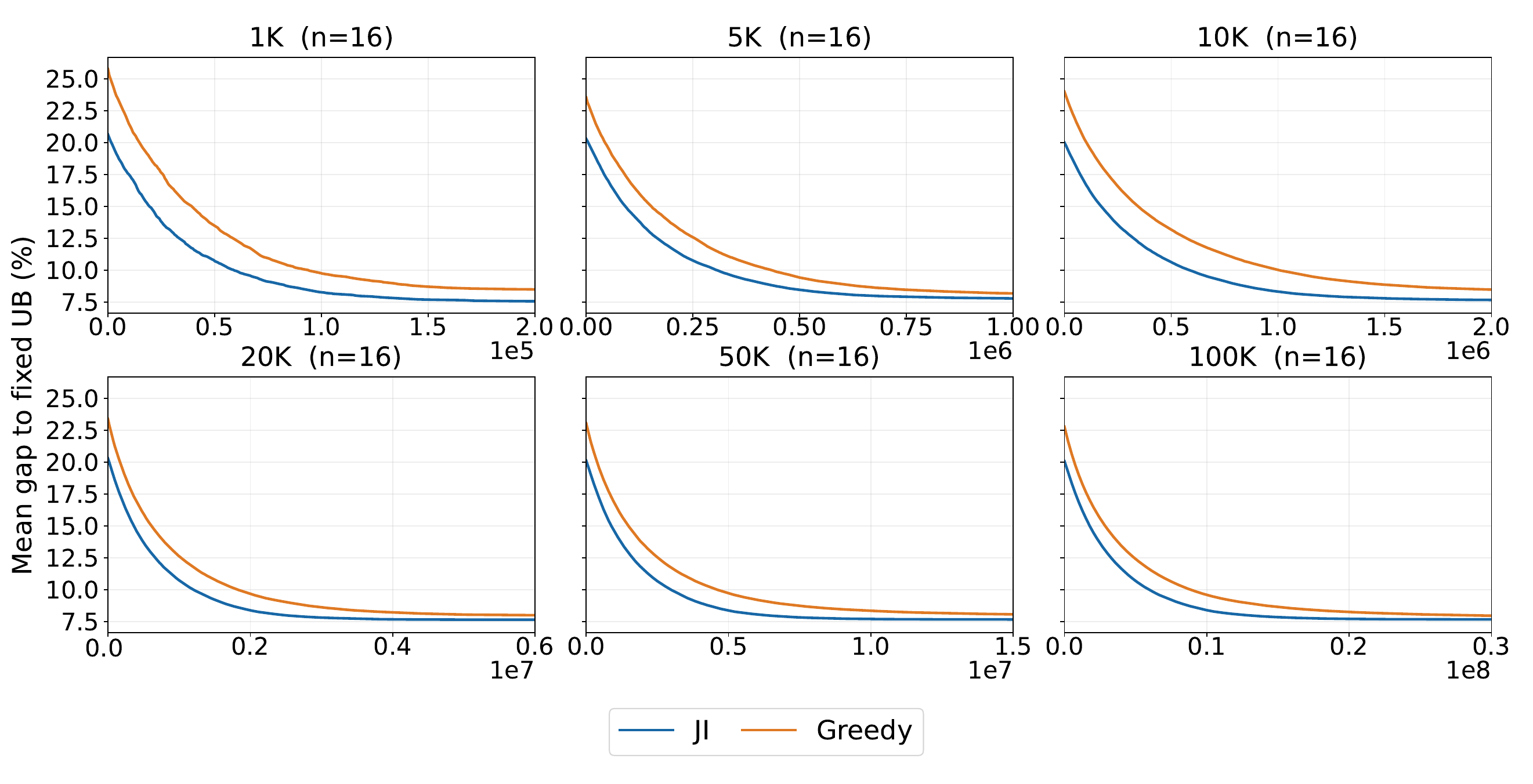}
\caption{GAP convergence trajectories during 2-opt refinement with greedy and JI initializations across six TSP scales.}
\label{fig:greedy_start_convergence}
\end{figure}

\paragraph{Results.}
Tables~\ref{tab:greedy_final} and~\ref{tab:search_budget}, together with Figure~\ref{fig:greedy_start_convergence}, 
show the refinement behavior of JI and greedy initialization.
JI consistently achieves lower final GAPs across all scales while using comparable or lower refinement time.
The convergence trajectories further demonstrate that JI improves more rapidly during early optimization and maintains a lower GAP throughout the search process.
Moreover, JI reaches the 10\% GAP target with fewer candidate evaluations, reducing the required search budget by 29--41\% across different scales.

\paragraph{Analysis.}
These results indicate that the improvement of JI comes from providing more optimization-friendly initial solutions rather than modifying the refinement algorithm itself.
Although greedy initialization produces feasible solutions with reasonable quality, it does not explicitly exploit the global structure of the solution space.
In contrast, JI identifies more promising regions through compact-space optimization, allowing downstream refinement to spend fewer evaluations exploring unfavorable regions.
This supports our hypothesis that improving initialization can reduce the computational burden of large-scale optimization by guiding search toward better solution regions.

\flushbottom

\end{document}